\documentclass{article}

\usepackage{neurips_2026}          % anonymous submission mode (no [final])
\nipsfinalcopy

\usepackage[utf8]{inputenc}
\usepackage[T1]{fontenc}
\usepackage{hyperref}
\usepackage{url}
\usepackage{booktabs}
\usepackage{amsmath}
\usepackage{graphicx}
\usepackage{subcaption}
\usepackage{longtable}
\usepackage{amssymb}
\usepackage{algorithm}
\usepackage{algpseudocode}
\usepackage{array} 
\usepackage[table]{xcolor}

\title{\Large {Generative Latent Priors for Interpreting \\ and Steering Protein Language Models}}

\author{%
  Daniel Hinjos\thanks{Equal contribution. Accepted at the Interpretability for Discovery (Interp4Discovery) Workshop at the 40th Conference on Neural Information Processing Systems (NeurIPS 2026)} \\ %Accepted at Interpretability for Discovery Workshop at NeurIPS 2026}  \\
  Barcelona Supercomputing Center \\
  Universitat Politècnica de Catalunya \\
  \texttt{daniel.hinjos@bsc.es} \\
  \And
  Lucía Urcelay\footnotemark[1] \\
  Centre for Genomic Regulation \\
  \texttt{lucia.urcelay@crg.eu} \\
}

\begin{document}
\maketitle
\renewcommand{\thefootnote}{}
\renewcommand{\thefootnote}{\arabic{footnote}}

\begin{abstract}
Dominant methods for interpreting large models impose strong structural assumptions. Generative Latent Priors (GLP) are a complementary approach: a diffusion
model that learns the distribution of a frozen model's activations directly, imposing no
assumptions. We port GLP to ESM-2 activations (UniRef50) and evaluate it
at two host scales (ESM-2-8M and -650M), both as a \textit{lens} for protein representations and
as a \textit{prior} for editing them. Generation quality scales smoothly with depth and compute. As a
\emph{lens}, single GLP ``meta-neurons'' are the strongest probes of discrete
Swiss-Prot concepts at both hosts, beating InterPLM's released SAE features, and lead on
continuous properties at 8M though not at 650M; falsification tests show this is mostly concentration, not new information. As an \emph{editing prior}, the on-manifold
correction makes DiffMean steering of three biochemical properties effective and restores on-manifoldness, but only in isolation: re-embedding the sequences actually produced after several rounds of decoding, the structural plausibility is compromised. This is mainly reproduced in additional SAE-based clamping structural steering experiments, where only Helix motif succeeds. The naturalness failure
is traced to the rewriting coverage and the decoder rather than the prior, highlighting what the method must clear
before it can serve for both \emph{discovery-by-editing} and \emph{discovery-by-exploration}. Trained GLPs are released on \href{https://huggingface.co/collections/HPAI-BSC/glprot}{huggingface.co/glprot}
 
\end{abstract}
\section{Introduction}

While protein language models (pLMs) have transformed sequence modeling in biotechnology and drug discovery, interpreting their internal representations remains a fundamental challenge. In the activation space, the dominating mechanistic interpretability methods rely on sparse
dictionary learning, training sparse autoencoders (SAEs) on ESM-2 \cite{lin2023esm2} activations and recovering features aligned with Swiss-Prot annotations \cite{simon2024interplm,gujral2025sparse,adams2025mechanistic}. However, these methods impose strong structural assumptions on the latent space. The Generative Latent Prior (GLP) \cite{luo2026glp} is a complementary \emph{generative} approach recently introduced for LLMs:
rather than decomposing activations into a sparse dictionary, a diffusion model is trained on a
frozen host's activations, learning their distribution directly with no linearity or sparsity
assumption. The learnt manifold serves as a tool for inspecting the latent space and moving through it, enabling navigation and control of the generations. 

\begin{figure}[H]
\centering
\includegraphics[width=1\textwidth]{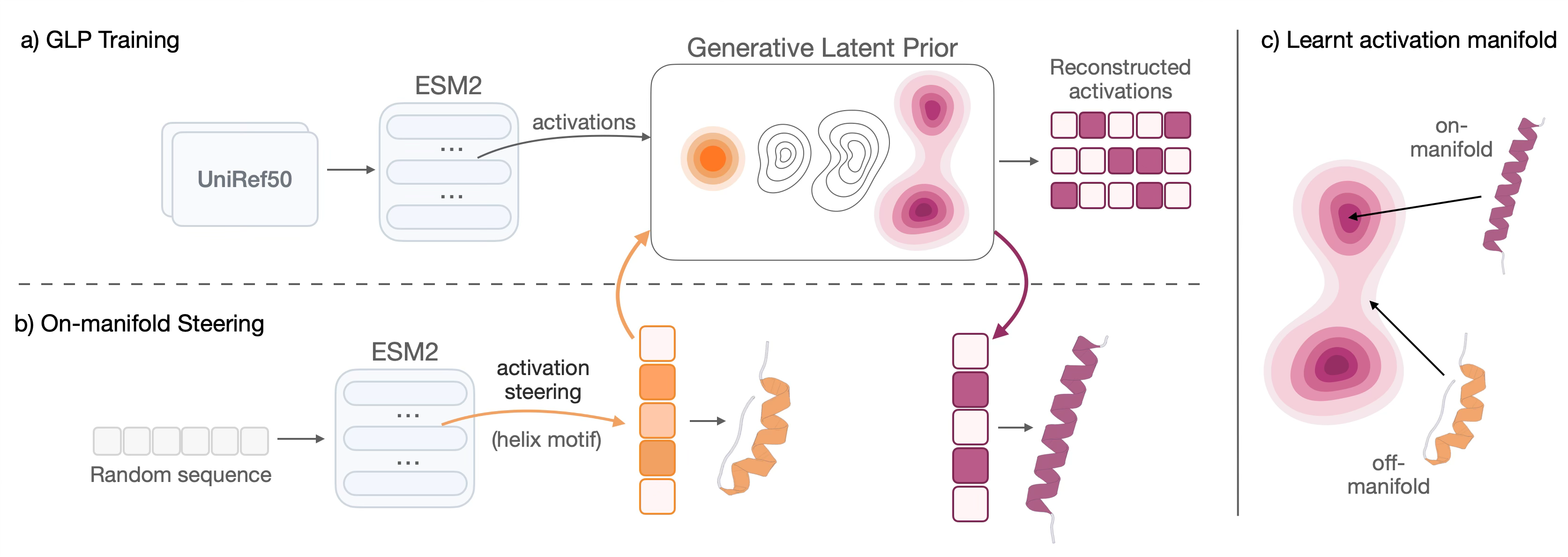}
\caption{Pipeline overview. (a) GLP trains by flow matching on ESM-2 activations from UniRef50
to learn their manifold. (b) Steering pushes an activation off that manifold toward a target
property. (c) GLP correction re-noises and denoises the edit back onto the manifold.}
\label{fig:pipeline}
\end{figure}

This is particularly relevant for activation steering \cite{turner2023actadd,rimsky2024caa}, a control approach based on editing model behavior, typically by adding a fixed linear direction to its activations. Large or naive edits can push them off the distribution the host was trained on, yielding representations the
host never produces. In this setting, GLP can serve as a corrector for steered activations, by partially noising and denoising
them back onto the learnt manifold, keeping the intended shift while restoring them to realistic
activations.

We port GLPs to pLMs and evaluate them as a reading lens and as an editing prior. Our contributions
are: (i) flow-matching GLPs trained on ESM-2-8M and ESM-2-650M activations, whose generation quality scales
smoothly with depth and compute; (ii) a single-unit probing analysis in which GLP meta-neurons
are the strongest probes of discrete Swiss-Prot concepts at both hosts, beating InterPLM
\cite{simon2024interplm} features, with falsification tests showing that this advantage comes mostly from concentration; and (iii) an evaluation of GLP as an on-manifold editing prior for correction under two steering
mechanisms: DiffMean directions \cite{wu2025axbench,huang2025diffmean} for three biochemical properties and SAE feature clamping for structural motifs. Together, these contributions depict GLP's role and limitations for
control and discovery in pLMs; Fig.~\ref{fig:pipeline} overviews the pipeline.

\section{Related Work}
\label{sec:relwork}

\paragraph{Activation modeling for pLMs.}
InterPLM \cite{simon2024interplm} matches SAE features against Swiss-Prot and
reports per-concept agreement; later work extends this to protein-level
representations and GO enrichment \cite{gujral2025sparse}, to a taxonomy of
features by activation pattern and family specificity \cite{adams2025mechanistic,villegas2025plmsae}. All of it
inherits the pathologies of sparse dictionary learning: a structured
reconstruction residual \cite{engels2024dark}, feature splitting and absorption
\cite{chanin2024absorption}, a granularity set by dictionary size rather than by
the data \cite{leask2025}, and dictionaries that differ across random seeds
\cite{paulo2025}. Sparsity and linearity are imposed, which leaves
open what a representation of the same activations fitted without those assumptions would isolate.

\paragraph{Activation steering for pLMs.}
Steering edits a model at inference time by adding a direction to its hidden
states, leaving the weights untouched. A direction can come from different approaches, such as the contrast between two sets of inputs \cite{turner2023actadd,rimsky2024caa,wu2025axbench} or from a
single unit of a sparse dictionary clamped at the positions where it fires
\cite{templeton2024scaling}. Both routes transfer to pLMs, where the contrast is
grounded in sequence-function rather than in linguistic opposition
\cite{huang2025diffmean} and the clamped latent in a structural or functional
annotation \cite{villegas2025plmsae}. We test both, on the three biochemical
properties studied in \cite{banerjee2025labelling} (hydrophobicity, instability, charge) and on structural motifs. These edits share a limitation: the resulting activation can land in a region the model never visits, hindering generation even while gaining the desired property. \cite{zhang2026offmanifold} documents this in guided pLMs and
responds by discarding affected candidates under a statistical density prior; we instead
correct the activation itself.

\section{Methods}
\subsection{Generative Latent Priors on ESM-2 activations}

\textbf{Substrate.} ESM-2-8M and ESM-2-650M ($d{=}320$ and $d{=}1280$, respectively) are frozen throughout. We pass UniRef50 sequences and read the residual stream after layers 4 and 18 respectively, both being layers InterPLM released SAEs for, so that both methods are comparable
(Appendix~\ref{app:pipeline}). We train GLPs on 100M (8M/650M) and 500M (8M)  activations from 5M random proteins. All
results below use the 24-block model.

\textbf{Architecture.} Let $h \in \mathbb{R}^{d}$ be one standardised per-residue activation
from that stream, drawn from an unknown distribution $p(h)$. A GLP is a flow-matching model of
$p(h)$: it learns to map Gaussian noise into a realistic on-manifold activation. Training interpolates a
real $h$ toward noise $\epsilon$ by a random amount $t$, with $\epsilon \sim \mathcal{N}(0,I)$, $t \in [0,1]$, giving $h_t = (1-t)h + t\epsilon$. As $t$ increases, $h_t$ moves along this line with
constant velocity $\frac{d h_t}{dt} = \epsilon - h$, and a network $u_\theta$ is trained to
predict this velocity target from $h_t$ and $t$ alone,
\begin{equation}
\mathcal{L}(\theta) = \mathbb{E}_{h,\epsilon,t}\big\lVert u_\theta(h_t,t) - (\epsilon - h) \big\rVert^2 ,
\label{eq:fm}
\end{equation}

so integrating $-u_\theta$ from $t{=}1$ down to $t{=}0$ removes noise. Started from pure noise,
this generates a new activation; started from an edited activation partly corrupted to an intermediate
level $t_0 < 1$, it pulls it back onto the activation manifold. Here $u_\theta$ is a stack of SwiGLU MLP blocks, depth 3--24; its per-block gated activations,
read at a fixed noise level $t{=}\tau$, are the \emph{meta-neurons} we probe. %% antes t=u, choca con u_theta
Full training hyperparameters are in Appendix~\ref{app:repro}.

\textbf{Generation quality.} We report the Fréchet Distance (FD) between generated and held-out
real activations, alongside the FD between two disjoint real splits, which is the best value
achievable at our sample size (Appendix~\ref{app:scaling}).

\subsection{Concept probing: what do meta-neurons isolate?}
\label{sec:method-probing}

We study whether GLP's meta-neurons acting as single units isolate biological and biochemical concepts. The concept set is frozen before any probing run and
built without reference to any GLP output: starting from the annotations that
pass InterPLM's frequency filter, we
discard residue-identity one-hots, which are trivially decodable, and
annotations without enough positive residues in the test split, leaving 110
discrete residue-level Swiss-Prot annotations and 10 continuous whole-sequence
properties. We compare four representations: GLP meta-neurons, InterPLM's SAE features, ESM's own
feedforward neurons, and the raw residual stream. Probes are trained on half the Swiss-Prot evaluation proteins (inner 5-fold CV selects the unit
and $\ell_2$ strength, giving the \emph{validation} AUC) and scored on the held-out half,
which is also excluded by accession from GLP's training activations (Appendix~\ref{app:pipeline}).

For \textbf{discrete concepts} we probe at residue level in three steps: (1) on the train split,
shortlist units whose activations best separate concept-containing (positive) from
concept-lacking (negative) residues; (2) fit each as a 1-D logistic probe and keep the highest
\emph{validation} AUC; (3) report that unit's AUC on a held-out \emph{test} split. Negatives come
from the \emph{same} proteins as positives, so a unit can only win by recognising the concept, not protein
identity.

As a falsification test, we verify whether each winning meta-neuron is linearly recoverable from the
full 10{,}240-feature SAE dictionary. Furthermore, we test whether a joint probe over the SAE's
top-$k$ features ($k{\le}20$)  matches the winning meta-neuron score on the concept. Together
the two tests separate \emph{concentration} from content: a dictionary that spans the meta-neuron's
direction yet needs $k$ features to exploit it carries the same concept information in a less
concentrated form
(Appendix~\ref{app:tables}).

For \textbf{continuous concepts}, we probe ten protein-level properties on mean-pooled activations, shortlisting units by the absolute Pearson correlation between their pooled activation and the computed property, and fitting a linear rather than logistic probe: GRAVY (hydrophobicity), molecular weight, isoelectric point, aromaticity, instability index, charge at pH 7, aliphatic index, helix fraction, turn fraction and sheet fraction. Each label is computed directly from the amino acid sequence. In this case, we report test Spearman $\rho$ instead of AUC, since a continuous property has no positive/negative split to define.

\subsection{On-manifold steering}
\label{sec:method-steering}

Steering edits typically rely on adding a fixed linear direction to an activation in order to bias generation towards a desired behavior. However, the edited activation is not tied to the host distribution, so it can land outside of the learnt manifold, degrading generations. A GLP that has learned the distribution of real activations offers a way to address this: project the
edit back onto the manifold before decoding. This raises some questions: does the
correction restore \emph{on-manifold-ness}, by the GLP's own diffusion loss, while increasing the property gain?
And, does an on-manifold correction decode into a structurally plausible
protein, measured by ESMFold pTM \cite{lin2023esm2}. We evaluate both under two edit types sharing a mechanism (Algorithm~\ref{alg:steer}): a masked-Gibbs resampling \cite{wang2019bert,johnson2021gibbs} with the correction applied at every step or no correction at all (``naive''). Reported experiments are on
ESM-2-8M/500M.

\textbf{Property steering.} Following Huang et al. \cite{huang2025diffmean},
we compute the steering direction $v$ as the mean-activation difference between the top and bottom UniRef50
quartiles of a biological property.
We scale $v$ to the mean activation norm $\overline{\lVert h\rVert}$,
$\hat v = \overline{\lVert h\rVert}\,v/\lVert v\rVert$, then add it to $h$ and rescale to
$\lVert h\rVert$, $\tilde h = \lVert h\rVert\,(h + \alpha\hat v)/\lVert h + \alpha\hat v\rVert$,
so the strength $\alpha$ varies direction alone (Appendix~\ref{app:steer-estimand}). We randomly sample and steer 10\% of residues without replacement for N=30 rounds. This results in substantial residue coverage, so we expect foldability to be compromised. Properties are scored directly from the amino-acid sequence: GRAVY \cite{kyte1982gravy}, net charge at pH~7 divided by length,
and the Guruprasad instability index \cite{guruprasad1990instability}.

\begin{figure}[H]
\centering
\includegraphics[width=1\textwidth]{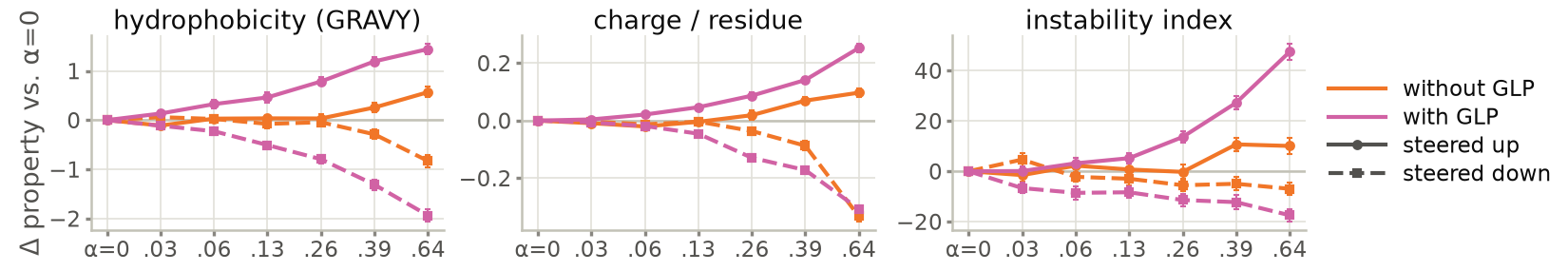}
\caption{Property steering naive vs GLP correction, both directions. Change vs.\ $\alpha{=}0$}
\label{fig:steer-effect-arms}
\end{figure}

We steer three properties (hydrophobicity, charge, instability
\cite{banerjee2025labelling}) in two directions: ``up'' from held-out low-scoring proteins and
``down'' from high-scoring ones. Each property uses 40 starting proteins per direction and 7
steering strengths ($\alpha$), with and without the correction, for 1{,}120 generations. We report the divergence, the effect of steering up minus the effect of steering down, in Appendix~\ref{app:steer-estimand}.
A genuine direction moves the two in opposite ways, so the divergence is large; an artifact of
the intervention itself shifts both the same way, so it cancels out. When reporting the up and down arms separately (Figure \ref{fig:steer-effect-arms}), we subtract the $\alpha{=}0$ score, since
the GLP correction alone already shifts the property (Appendix \ref{app:steer-drift}).

\textbf{Structural steering} instead overwrites one SAE
latent to a clamp value, $z_k \!\leftarrow\! c$, and reconstructs the activation, residual included, $\tilde h = \mathrm{Dec}(z') + h - \mathrm{Dec}(z)$. Following
\cite{villegas2025plmsae} we start from short random sequences, clamp and resample the
whole sequence. We do 100 independent rounds, keeping the round maximising $z_k$. A structural concept the SAE isolates as
roughly one feature can be pushed directly: for Helix, Beta strand, Coiled coil and
Zinc finger we clamp the feature with the highest feature--concept F1 against
Swiss-Prot labels. Success is a concept hit in the finished sequence, so we fold with
ESMFold and detect Helix and Beta strand with DSSP, Coiled coil with
\texttt{pepcoil}, and Zinc finger with a PROSITE \cite{sigrist2026prosite} scan
corroborated by a Pfam zf-C2H2 HMM alignment \cite{sathyaseelan2023znfprot}  (Appendix~\ref{app:steer-sae-clamp}).

\begin{table}[h]
\centering
\caption{1-D probing: mean test AUC or Spearman $\rho$ [95\% bootstrap CI]. Bold: highest per column; $^{*}$: $p{<}0.05$ vs.\ all baselines (two-sided paired Wilcoxon).}
\label{tab:probing}
\resizebox{\textwidth}{!}{%
\begin{tabular}{lcccc}
\toprule
 & \multicolumn{2}{c}{Discrete AUC} & \multicolumn{2}{c}{Continuous $\rho$} \\
\cmidrule(lr){2-3}\cmidrule(lr){4-5}
Representation & 8M  & 650M  & 8M  & 650M \\
\midrule
GLP  & \textbf{0.903}$^{*}$ [0.887, 0.918] & \textbf{0.932}$^{*}$ [0.917, 0.946] & \textbf{0.669}$^{*}$ [0.617, 0.725] & 0.555 [0.498, 0.613] \\
SAE \cite{simon2024interplm}      & 0.845 [0.826, 0.865]          & 0.872 [0.850, 0.893]          & 0.602 [0.557, 0.650]          & \textbf{0.569} [0.498, 0.648] \\
ESM FFN     & 0.877 [0.859, 0.894]          & 0.850 [0.829, 0.871]          & 0.563 [0.525, 0.600]          & 0.373 [0.325, 0.420] \\
Res. Stream & 0.833 [0.815, 0.850]          & 0.872 [0.852, 0.891]          & 0.478 [0.413, 0.549]          & 0.373 [0.317, 0.424] \\
\bottomrule
\end{tabular}%
}
\end{table}

\iffalse
\begin{table}[h]
\centering
\caption{Effect of the GLP correction on steering efficacy: up $-$ down divergence in units of
each property's natural spread, at an edit of 26\% of the mean activation norm ($\alpha{=}0.257$).}
\label{tab:steer-div}
\begin{tabular}{lrrrr}
\toprule
property & without GLP & $p$ & with GLP & $p$ \\
\midrule
hydrophobicity (GRAVY) & $+0.17$ & $0.44$ & $+3.46$ & $2{\times}10^{-13}$ \\
charge / residue       & $+0.92$ & $0.02$ & $+3.60$ & $2{\times}10^{-14}$ \\
instability index      & $+0.34$ & $0.19$ & $+1.59$ & $7{\times}10^{-9}$ \\
\bottomrule
\end{tabular}
\end{table}
\fi

\section{Results}

\subsection{Generation quality scales with compute}
\label{sec:gen-quality}

GLP trains stably, converging well below the trivial-predictor floor. With enough sampler steps,
its samples match the real activation distribution on its top two principal components
(Fig.~\ref{fig:pca}). Two properties emerge across a grid of four depths and two
data scales. First, the diffusion loss follows a compute trend: plotted against $C{=}6ND$
\cite{kaplan2020}, all eight runs collapse onto one curve (Fig.~\ref{fig:scaling}a), so the
loss is set by the total compute budget, not by how it is split between depth (model size) and
data. Second, Fréchet Distance to real activations improves with depth, with the gain
concentrated at $12$--$24$ blocks,
and this depth-scaling reproduces on ESM-2-650M. Absolute FD scales with the width of the activation space, so we divide each host's FD by its own real-versus-real floor (the FD between two disjoint samples of real activations) giving a
dimensionless ratio comparable across hosts (Fig.~\ref{fig:scaling}b).

\subsection{Concept information concentrates in single meta-neurons}
\label{sec:probing-results}

\textbf{GLP's meta-neurons are the best unit probes on most concepts.} Table~\ref{tab:probing}
reports mean test AUC over the 110 qualifying Swiss-Prot concepts and mean test Spearman
$\rho$ over 10 continuous properties. We test each GLP-vs-baseline margin as a matched pair and a paired
bootstrap over these differences (10,000 resamples) gives the 95\% CIs. At 8M, GLP meta-neurons lead every baseline on
both. The margins are small in absolute terms but consistent concept by concept, which is
what the paired intervals in the table reflect: a meta-neuron is the better single unit on
104 of 110 concepts against the SAE, 92 against ESM's feedforward neurons and 106 against
the raw residual stream, and on 8, 8 and 10 of the 10 continuous properties, respectively
(Appendix~\ref{app:tables}).

\textbf{The discrete lead holds at scale.} The same 110 concepts probed on GLP-ESM-2-650M result in the same findings: a meta-neuron isolates concepts better
than every baseline, winning 105 concepts against the SAE, 109 against the feedforward
neurons and 102 against the residual stream (all paired $p{<}10^{-17}$). Absolute AUCs rise
with the host, but the GLP-over-SAE advantage margin barely moves
($+0.048$ at 650M against $+0.058$ at 8M). However, the lead is specific to discrete concepts, as on the ten continuous properties the SAE nominally edges it with overlapping CIs.

\textbf{Meta-neurons concentrate what the sparse dictionary spreads.} Winning on a single
unit could just mean GLP found a better rotation of a subspace the SAE already spans, and two tests say it largely does. Each winning meta-neuron is mostly a linear
read-out of the SAE dictionary (median $R^2 = 0.76$ at 8M, $0.73$ at 650M), and a joint probe
over the 20 most concept-selective SAE features reaches its test AUC for the large majority
of concepts where GLP wins (94 of 104 at 8M, 101 of 105 at 650M). What one meta-neuron reads, the dictionary holds too, but spread over several atoms.

For a few concepts the joint probe never catches the meta-neuron: 10 of 104 at 8M, 4 of 105
at 650M. These are near-misses rather than failures. The gap is larger on one concept per host, but not on the same concept at the
two hosts (Appendix~\ref{app:tables}), so we do not read it as a concept the dictionary cannot
represent. Thus, the single-unit advantage seems to be concentration.

\subsection{Correction succeeds but naturalness is compromised by decoding}
\label{sec:steering-results}

\textbf{On-manifold correction enables effective property steering.} The naive arm (no
correction) barely separates the up and down directions except at the strongest
perturbation, whereas the correction separates them cleanly on all three properties
across most of the grid (Figure~\ref{fig:steer-effect-arms}; full dose-divergence response in
Appendix~\ref{app:steer-estimand}). Steering does move the activation off the manifold, more so at larger $\alpha$, and a single
correction pass restores its typicality, increasing the property gain (Appendix~\ref{app:steer-sdedit}).  This buys a gentler intervention: the corrected arm
already moves the property at perturbations the naive arm cannot resolve at all, so the
property can be changed with a far smaller edit.

\newpage

\textbf{Structural steering is concept-dependent and locality-limited.} Clamping a single SAE
feature toward four structural concepts does not reproduce the property steering result, as only Helix
improves consistently. An apparent Coiled coil gain is simply helix formation scored as hits, Beta strand falls under correction and Zinc finger yields no motif (though a continuous HMM score shows a shift
toward zf-C2H2 character). This is what a per-residue intervention can reach: while a helix is set by local propensities within a short window, the rest need residues far apart in sequence to pair, harder to generate. Feature selection is also indirect: we select the clamp target by detection F1,
assuming the best identifier is the best controller, yet the strongest detector (F1 $0.907$ for Zinc Finger) never
finds a hit, so better, more sophisticated selection methods might reach concepts this rule cannot. Foldability is compromised
throughout, following the same pattern as property steering. All results are reported in Appendix~\ref{app:steer-sae-clamp}.

\textbf{Rewriting, not the prior, limits foldability and on-manifold restoration.} Previous steering studies report concept efficacy but not folding \cite{villegas2025plmsae,huang2025diffmean}; we fold every generation. The masked-Gibbs decoding loop drives every starting protein to a common, mediocre fold confidence regardless of where it started (Table~\ref{tab:steer-decode6}, Appendix~\ref{app:steer-protocol}), and both arms land there together, so the correction neither helps nor hurts foldability. What costs structure is how much of
the protein gets rewritten (Appendix~\ref{app:steer-decoders}). Typicality is recovered but not retained: a correction pass restores GLP's diffusion loss on an isolated steered activation (Appendix~\ref{app:steer-sdedit}), yet, after re-embedding the finished sequences, the corrected arm is no more on-manifold than the naive one, under either steering method. A prior over activations buys control over the property, not over the sequence that carries it. What this leaves open is the granularity of the intervention, which our methodology, a uniform push on every resampled residue, does not address.

\section{Conclusions}
\label{sec:conclusions}

In this study we port GLPs to pLMs, and evaluate them as
a lens for reading pLM's internal representations and as a prior for editing them. As a lens, a single
GLP meta-neuron out-probes mostly every baseline (the SAE, ESM's own feedforward units,
the raw residual stream) at both ESM-2-8M and -650M, with some exceptions, but our falsification tests read most of this as \emph{concentration}, not new information. GLPs are still a faithful, better-reading lens. As
an editing prior, the correction is mostly effective for steering across biochemical properties and some restrained structural motifs, such as Helix. However, although the on-manifold edit has great effect in isolation, neither post-correction foldability nor on-manifoldness survive the decoding loop in the implemented masked-Gibbs algorithm (Section~\ref{sec:steering-results}). This is a structural limit on
\emph{discovery-by-editing} when using a continuous prior to steer protein sequence generations.

Our study is bounded to training on one hook layer per host, correlational probing on a domain/motif-heavy
pool and steering restricted to one layer and the masked positions. Future work splits along
the same two roles explored. On the lens side, meta-neurons are read only by 1-D probing against curated annotations; automated
interpretation of individual units and of the manifold's local geometry could
characterise what they encode beyond them. On the editing side, naturalness may improve with a better decoding
algorithm (beam search already folds better per edit than masked-Gibbs,
Appendix~\ref{app:steer-decoders}), or by adjusting the masked-Gibbs loop for less, more granular and guided coverage. Another approach falls into avoiding decoding, considering a host whose output stays continuous throughout (structural or latent
diffusion), or an edit constrained inside the sampler itself (either clamping a meta-neuron to a target
value or gradient-guiding the denoising trajectory toward one) so the result is on-manifold by construction. Here, the correction pulled toward high density for control and discovery-by-editing; the same density estimate could instead detect and hold activations at the boundary of typicality, turning it into a tool for actual novel discovery.

% Acknowledgements omitted for anonymous submission (deanonymizing); restore for camera-ready.
% \medskip
\noindent\textbf{Acknowledgements.} This work was supported by Daniel Hinjos García's fellowship within the ``Generación D'' initiative, Red.es, Ministerio para la Transformación Digital y de la Función Pública, for talent attraction (C005/24-ED CV1), funded by the European Union
 NextGenerationEU funds, through PRTR. We acknowledge the computational resources provided by the MareNostrum 5 supercomputer.

\bibliographystyle{plain} 
\bibliography{references}

@article{lin2023esm2,
  title   = {Evolutionary-scale prediction of atomic-level protein structure with a language model},
  author  = {Lin, Zeming and Akin, Halil and Rao, Roshan and Hie, Brian and Zhu, Zhongkai and Lu, Wenting and Smetanin, Nikita and Verkuil, Robert and Kabeli, Ori and Shmueli, Yaniv and dos Santos Costa, Allan and Fazel-Zarandi, Maryam and Sercu, Tom and Candido, Salvatore and Rives, Alexander},
  journal = {Science},
  volume  = {379},
  number  = {6637},
  pages   = {1123--1130},
  year    = {2023},
  doi     = {10.1126/science.ade2574}
}

@article{simon2024interplm,
  title   = {{InterPLM}: discovering interpretable features in protein language models via sparse autoencoders},
  author  = {Simon, Elana and Zou, James},
  journal = {Nature Methods},
  volume  = {22},
  number  = {10},
  pages   = {2107--2117},
  year    = {2025},
  doi     = {10.1038/s41592-025-02836-7}
}

@article{luo2026glp,
  title   = {Learning a generative meta-model of {LLM} activations},
  author  = {Luo, Grace and Feng, Jiahai and Darrell, Trevor and Radford, Alec and Steinhardt, Jacob},
  journal = {arXiv preprint arXiv:2602.06964},
  year    = {2026},
  url     = {https://arxiv.org/abs/2602.06964}
}

@article{kaplan2020,
  title   = {Scaling laws for neural language models},
  author  = {Kaplan, Jared and McCandlish, Sam and Henighan, Tom and Brown, Tom B. and Chess, Benjamin and Child, Rewon and Gray, Scott and Radford, Alec and Wu, Jeffrey and Amodei, Dario},
  journal = {arXiv preprint arXiv:2001.08361},
  year    = {2020},
  url     = {https://arxiv.org/abs/2001.08361}
}

@article{villegas2025plmsae,
  title   = {Interpreting and steering protein language models through sparse autoencoders},
  author  = {{Villegas Garc{\'i}a}, Edith Natalia and Ansuini, Alessio},
  journal = {arXiv preprint arXiv:2502.09135},
  year    = {2025},
  url     = {https://arxiv.org/abs/2502.09135}
}

@article{huang2025diffmean,
  title   = {Steering protein language models},
  author  = {Huang, Long-Kai and Zhu, Rongyi and He, Bing and Yao, Jianhua},
  journal = {arXiv preprint arXiv:2509.07983},
  year    = {2025},
  url     = {https://arxiv.org/abs/2509.07983}
}

@inproceedings{banerjee2025labelling,
  title     = {Automated neuron labelling enables generative steering and interpretability in protein language models},
  author    = {Banerjee, Arjun and Martinez, David and Dang, Camille and Tam, Ethan},
  booktitle = {ICML Workshop on Generative AI for Biology},
  series    = {PMLR},
  volume    = {267},
  year      = {2025},
  note      = {arXiv:2507.06458}
}

@inproceedings{chang2022maskgit,
  title     = {{MaskGIT}: Masked generative image transformer},
  author    = {Chang, Huiwen and Zhang, Han and Jiang, Lu and Liu, Ce and Freeman, William T.},
  booktitle = {IEEE/CVF Conference on Computer Vision and Pattern Recognition (CVPR)},
  year      = {2022}
}

@article{beamsearch2026plm,
  title   = {How to make the most of your masked language model for protein engineering},
  author  = {McCarter, Calvin and others},
  journal = {arXiv preprint arXiv:2603.10302},
  year    = {2026},
  url     = {https://arxiv.org/abs/2603.10302}
}

@article{sathyaseelan2023znfprot,
  title   = {Sequence patterns and {HMM} profiles to predict proteome wide zinc finger motifs},
  author  = {Sathyaseelan, Chakkarai and Patro, L. Ponoop Prasad and Rathinavelan, Thenmalarchelvi},
  journal = {Pattern Recognition},
  volume  = {135},
  pages   = {109134},
  year    = {2023},
  doi     = {10.1016/j.patcog.2022.109134}
}

@article{gujral2025sparse,
  title   = {Sparse autoencoders uncover biologically interpretable features in protein language model representations},
  author  = {Gujral, Onkar and Bafna, Mihir and Alm, Eric and Berger, Bonnie},
  journal = {Proceedings of the National Academy of Sciences},
  volume  = {122},
  number  = {34},
  pages   = {e2506316122},
  year    = {2025},
  doi     = {10.1073/pnas.2506316122}
}

@article{adams2025mechanistic,
  title   = {From mechanistic interpretability to mechanistic biology: training, evaluating, and interpreting sparse autoencoders on protein language models},
  author  = {Adams, Etowah and Bai, Liam and Lee, Minji and Yu, Yiyang and AlQuraishi, Mohammed},
  journal = {bioRxiv},
  year    = {2025},
  doi     = {10.1101/2025.02.06.636901}
}

@article{sigrist2026prosite,
  title   = {The {PROSITE} database for protein families, domains, and sites},
  author  = {Sigrist, Christian J. A. and Cuche, B{\'e}atrice A. and de Castro, Edouard and Coudert, Elisabeth and Redaschi, Nicole and Bridge, Alan},
  journal = {Nucleic Acids Research},
  volume  = {54},
  number  = {D1},
  pages   = {D451--D458},
  year    = {2026},
  doi     = {10.1093/nar/gkaf1188}
}

@article{zhang2026offmanifold,
  title        = {Off-Manifold Collapse in Guided Protein Language Models},
  author       = {Zhang, Shuibai and Liu, Xinchi and Peng, Fred Zhangzhi and Yang, Zhihan and Wu, Shutong and Ma, Yingzi and Zhang, Jiawei},
  journal      = {arXiv preprint arXiv:2608.18597},
  year         = {2026},
  eprint       = {2608.18597},
  archivePrefix= {arXiv},
  primaryClass = {cs.LG}
}

@article{turner2023actadd,
  title        = {Activation Addition: Steering Language Models Without Optimization},
  author       = {Turner, Alexander Matt and Thiergart, Lisa and Leech, Gavin and Udell, David and Vazquez, Juan J. and Mini, Ulisse and MacDiarmid, Monte},
  journal      = {arXiv preprint arXiv:2308.10248},
  year         = {2023},
  eprint       = {2308.10248},
  archivePrefix= {arXiv},
  primaryClass = {cs.CL}
}

@inproceedings{rimsky2024caa,
  title     = {Steering {L}lama 2 via Contrastive Activation Addition},
  author    = {Rimsky, Nina and Gabrieli, Nick and Schulz, Julian and Tong, Meg and Hubinger, Evan and Turner, Alexander Matt},
  booktitle = {Proceedings of the 62nd Annual Meeting of the Association for Computational Linguistics (Volume 1: Long Papers)},
  pages     = {15504--15522},
  year      = {2024},
  address   = {Bangkok, Thailand},
  publisher = {Association for Computational Linguistics}
}

@article{engels2024dark,
  title        = {Decomposing the Dark Matter of Sparse Autoencoders},
  author       = {Engels, Joshua and Riggs, Logan and Tegmark, Max},
  journal      = {arXiv preprint arXiv:2410.14670},
  year         = {2024},
  eprint       = {2410.14670},
  archivePrefix= {arXiv},
  primaryClass = {cs.LG}
}

@article{chanin2024absorption,
  title        = {A is for Absorption: Studying Feature Splitting and Absorption in Sparse Autoencoders},
  author       = {Chanin, David and Wilken-Smith, James and Dulka, Tom{\'a}{\v{s}} and Bhatnagar, Hardik and Bloom, Joseph},
  journal      = {arXiv preprint arXiv:2409.14507},
  year         = {2024},
  eprint       = {2409.14507},
  archivePrefix= {arXiv},
  primaryClass = {cs.LG}
}

@inproceedings{leask2025,
  title     = {Sparse Autoencoders Do Not Find Canonical Units of Analysis},
  author    = {Leask, Patrick and Bussmann, Bart and Pearce, Michael T. and Bloom, Joseph Isaac and Tigges, Curt and Al Moubayed, Noura and Sharkey, Lee and Nanda, Neel},
  booktitle = {The Thirteenth International Conference on Learning Representations (ICLR)},
  year      = {2025},
  address   = {Singapore}
}

@article{paulo2025,
  title        = {Sparse Autoencoders Trained on the Same Data Learn Different Features},
  author       = {Paulo, Gon{\c{c}}alo and Belrose, Nora},
  journal      = {arXiv preprint arXiv:2501.16615},
  year         = {2025},
  eprint       = {2501.16615},
  archivePrefix= {arXiv},
  primaryClass = {cs.LG}
}

@article{templeton2024scaling,
  title   = {Scaling Monosemanticity: Extracting Interpretable Features from {C}laude 3 {S}onnet},
  author  = {Templeton, Adly and Conerly, Tom and Marcus, Jonathan and Lindsey, Jack and Bricken, Trenton and Chen, Brian and Pearce, Adam and Citro, Craig and Ameisen, Emmanuel and Jones, Andy and Cunningham, Hoagy and Turner, Nicholas L. and McDougall, Callum and MacDiarmid, Monte and Tamkin, Alex and Durmus, Esin and Hume, Tristan and Mosconi, Francesco and Freeman, C. Daniel and Sumers, Theodore R. and Rees, Edward and Batson, Joshua and Jermyn, Adam and Carter, Shan and Olah, Chris and Henighan, Tom},
  journal = {Transformer Circuits Thread},
  year    = {2024},
  url     = {https://transformer-circuits.pub/2024/scaling-monosemanticity/}
}

@inproceedings{wu2025axbench,
  title     = {{AxBench}: Steering {LLMs}? Even Simple Baselines Outperform Sparse Autoencoders},
  author    = {Wu, Zhengxuan and Arora, Aryaman and Geiger, Atticus and Wang, Zheng and Huang, Jing and Jurafsky, Dan and Manning, Christopher D. and Potts, Christopher},
  booktitle = {Proceedings of the 42nd International Conference on Machine Learning (ICML)},
  year      = {2025},
  eprint    = {2501.17148},
  archivePrefix = {arXiv}
}

@inproceedings{wang2019bert,
  title     = {{BERT} has a Mouth, and It Must Speak: {BERT} as a {Markov} Random Field Language Model},
  author    = {Wang, Alex and Cho, Kyunghyun},
  booktitle = {Proceedings of the Workshop on Methods for Optimizing and Evaluating Neural Language Generation},
  pages     = {30--36},
  year      = {2019}
}

@article{johnson2021gibbs,
  title   = {Generating novel protein sequences using {Gibbs} sampling of masked language models},
  author  = {Johnson, Sean R. and Monaco, Sarah and Massie, Kenneth and Syed, Zaid},
  journal = {bioRxiv},
  year    = {2021},
  doi     = {10.1101/2021.01.26.428322}
}

@article{kyte1982gravy,
  title   = {A simple method for displaying the hydropathic character of a protein},
  author  = {Kyte, Jack and Doolittle, Russell F.},
  journal = {Journal of Molecular Biology},
  volume  = {157},
  number  = {1},
  pages   = {105--132},
  year    = {1982}
}

@article{guruprasad1990instability,
  title   = {Correlation between stability of a protein and its dipeptide composition: a novel approach for predicting in vivo stability of a protein from its primary sequence},
  author  = {Guruprasad, Kunchur and Reddy, B. V. Bhaskara and Pandit, Madhusudan W.},
  journal = {Protein Engineering},
  volume  = {4},
  number  = {2},
  pages   = {155--161},
  year    = {1990}
}

\newpage

\appendix

\section{Scaling and generation quality}
\label{app:scaling}

The scaling grid crosses four GLP depths ($3$, $6$, $12$ and $24$ SwiGLU blocks) with two data
scales ($100$M and $500$M activations), giving eight models on the ESM-2-8M host, plus the four
depths at $100$M on ESM-2-650M (Table~\ref{tab:scaling}). 

\begin{figure}[H]
\centering
\begin{subfigure}[b]{0.55\textwidth}
  \includegraphics[width=\textwidth]{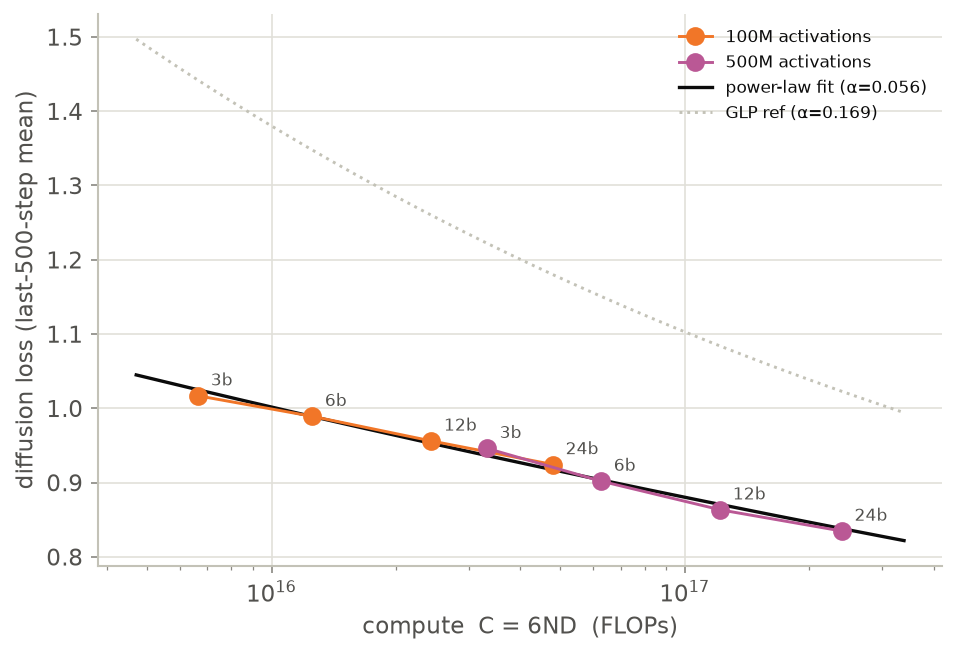}
  \caption{Compute scaling of the diffusion loss.}
\end{subfigure}\\[1em]
\begin{subfigure}[b]{0.55\textwidth}
  \includegraphics[width=\textwidth]{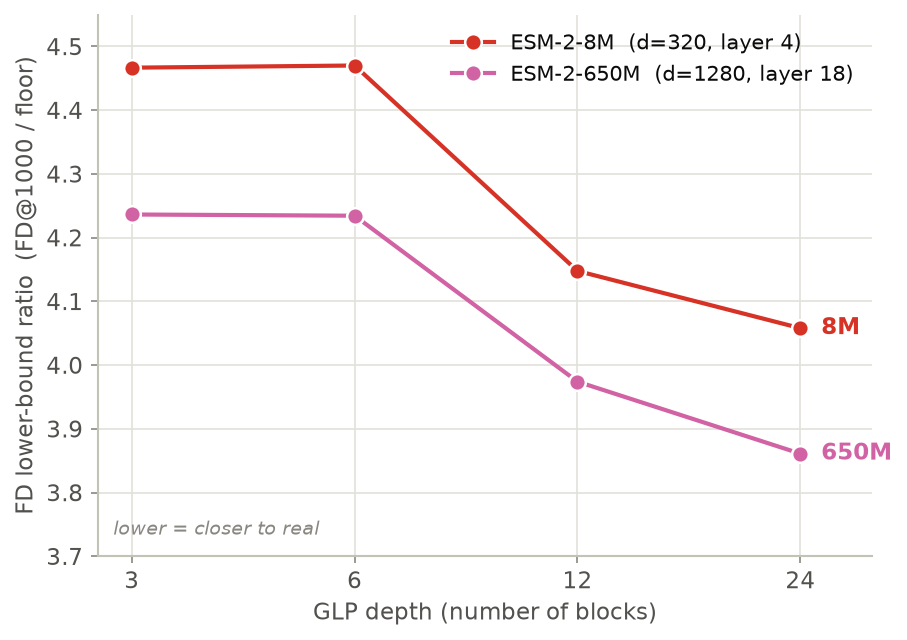}
  \caption{FD lower-bound ratio vs depth, both hosts.}
\end{subfigure}
\caption{Generation quality scales with compute. (a) Smoothed final loss of all eight
(depth $\times$ data) runs collapses onto one power law in $C{=}6ND$; our protein-activation
curve lies below the published LLM-activation reference (dotted). (b) FD lower-bound ratio
(FD@1000 / real-vs-real floor; lower is better) vs GLP depth, on matched 100M-activation grids}
\label{fig:scaling}
\end{figure}

Fig.~\ref{fig:scaling}a plots the final training loss of those eight ESM-2-8M-host runs against
compute $C = 6ND$ \cite{kaplan2020}, being $N$ the GLP's parameter count and $D$ the number of
activations seen. They collapse onto one curve: the loss is set by total compute, not
by how it is split between depth and data. Fitting $L(C) = E + A\,C^{-\alpha}$, where $E$ is the floor the loss would approach with
infinite compute, gives $\alpha \approx 0.056$, against $\alpha \approx 0.169$ reported by the
original GLP work on language-model activations; their loss falls faster with compute, but our
entire curve sits lower. Our compute range spans about one order of magnitude, too short to
estimate $E$, so we report only the slope and treat the fit as descriptive, rather than as a law.

\begin{figure}[t]
\centering
\includegraphics[width=0.98\textwidth]{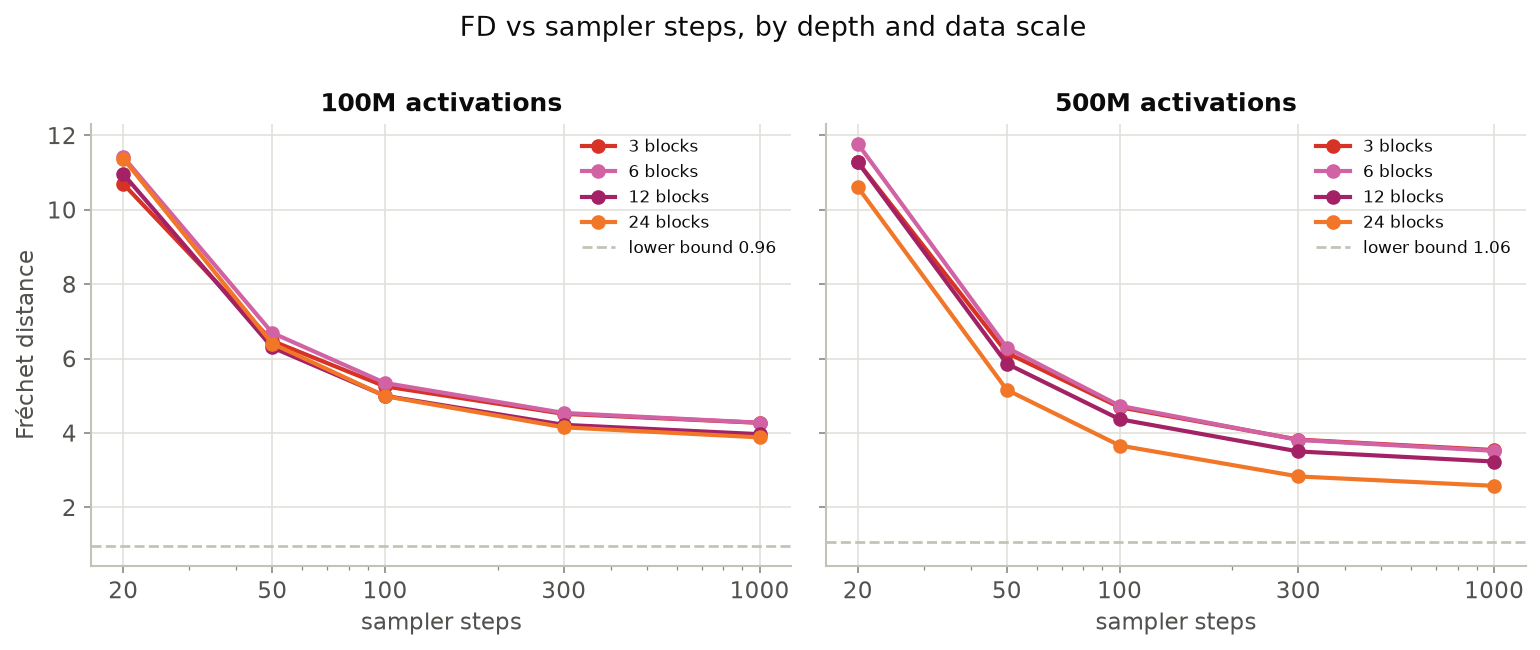}
\caption{Fréchet Distance vs.\ sampler steps, by GLP depth and data scale (shared
$y$-axis). FD falls with sampler steps everywhere. Depth lowers FD monotonically and the gain
is larger at $500$M than at $100$M: more data lets extra depth pay off. The shallowest cells
($3$ vs.\ $6$ blocks) are near-identical at both scales; separation appears at $12$--$24$
blocks. Dashed lines: real-vs-real lower bound.}
\label{fig:fdsteps}
\end{figure}

Fréchet Distance compares generated and real activations by their mean and covariance. Because
our generator is a diffusion sampler, FD at few steps mostly reports \emph{sampler} error rather
than model quality (Fig.~\ref{fig:fdsteps}), so every FD number is stated with its step count
(we use 1{,}000). We also report FD as a ratio to the FD(train, val) real-vs-real floor, the
smallest FD achievable from finite samples of the true distribution, which makes the number
comparable across hosts of different width.

\begin{table}[h]
\centering
\caption{Depth $\times$ data-scale grid, generation quality only. $N$ = parameters,
$C = 6ND$, FD at 1000 sampler steps; ratio is FD / real-vs-real lower bound; loss is the
mean of the last 500 training steps. The two hosts occupy different-dimensional activation
spaces ($d{=}320$ vs.\ $1280$), so absolute FD and loss are not comparable across host blocks;
only the FD ratio is. Bold: the model used for all probing and steering results.}
\label{tab:scaling}
\begin{tabular}{lrrrrrr}
\toprule
Data & Blocks & Params & $C = 6ND$ & FD@1000 & FD ratio & Loss \\
\midrule
\multicolumn{7}{l}{\emph{ESM-2-8M host ($d{=}320$)}} \\
100M & 3  & 11.1M & $6.7{\times}10^{15}$ & 4.27 & 4.47$\times$ & 1.017 \\
100M & 6  & 20.9M & $1.3{\times}10^{16}$ & 4.28 & 4.47$\times$ & 0.989 \\
100M & 12 & 40.6M & $2.4{\times}10^{16}$ & 3.97 & 4.15$\times$ & 0.956 \\
100M & 24 & 80.0M & $4.8{\times}10^{16}$ & 3.88 & 4.06$\times$ & 0.924 \\
\cmidrule(l){1-7}
500M & 3  & 11.1M & $3.3{\times}10^{16}$ & 3.54 & 3.33$\times$ & 0.946 \\
500M & 6  & 20.9M & $6.3{\times}10^{16}$ & 3.52 & 3.31$\times$ & 0.902 \\
500M & 12 & 40.6M & $1.2{\times}10^{17}$ & 3.23 & 3.04$\times$ & 0.863 \\
\textbf{500M} & \textbf{24} & \textbf{80.0M} & $2.4{\times}10^{17}$ & \textbf{2.58} & \textbf{2.42$\times$} & \textbf{0.835} \\
\midrule
\multicolumn{7}{l}{\emph{ESM-2-650M host ($d{=}1280$, layer 18)}} \\
100M & 3  & 177M  & $1.1{\times}10^{17}$ & 1346.8 & 4.24$\times$ & 1.038 \\
100M & 6  & 334M  & $2.0{\times}10^{17}$ & 1346.1 & 4.23$\times$ & 0.974 \\
100M & 12 & 649M  & $3.9{\times}10^{17}$ & 1263.4 & 3.97$\times$ & 0.919 \\
\textbf{100M} & \textbf{24} & \textbf{1.28B} & $7.7{\times}10^{17}$ & \textbf{1227.4} & \textbf{3.86$\times$} & \textbf{0.877} \\
\bottomrule
\end{tabular}
\end{table}

\begin{figure}[H]
\centering
\includegraphics[width=0.98\textwidth]{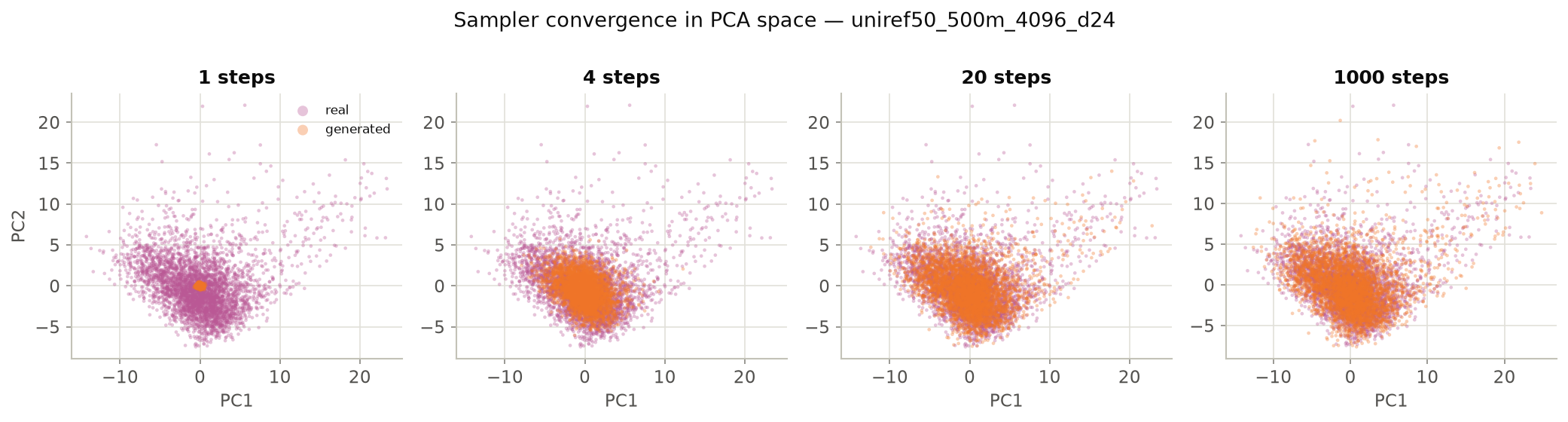}
\caption{Sampler convergence in PCA space (24-block, 500M): generated activations
(orange) refine from a collapsed blob at 1 step to matching the real distribution (purple)
at 1000 steps, on the real data's top-2 principal components. Inset: training loss, converging
below the trivial-predictor floor ($\approx 2.0$).}
\label{fig:pca}
\end{figure}

\newpage
\section{Data pipeline and reproducibility}
\label{app:data}

\subsection{Hook parity and the activation store}
\label{app:pipeline}

GLP is only a faithful model of what ESM-2 computes if it trains on exactly the tensor of
interest. We read \texttt{hidden\_states[4]} (the residual stream after transformer block 4)
with cls/eos tokens dropped, which is byte-identical to what InterPLM's released SAE consumed. Activations are extracted once per
data scale into a memory-mapped \texttt{float16} store with a fitted standardizer, and all depth
variants at that scale read the same store.

\textbf{Training data sampling.} The training pool is a one-pass sample of 5M sequences from the local UniRef50 FASTA, matching InterPLM's ``random
sequences from UniRef50''. The sampled sequences are embedded at the hooked layer in sampling order and their
per-residue activations streamed into the store until it reaches the target count, truncating
mid-sequence. Because the sampling order is random, the truncated store is an unbiased
per-residue subsample of the pool.

\textbf{Train/eval disjointness.} The Swiss-Prot proteins used for probing are almost all
present in UniRef50 too (${\sim}98\%$ overlap), so we drop every evaluation accession from GLP's
training sample. Otherwise, GLP would be fitted on the very activations it is later probed on.

\subsection{Training hyperparameters}
\label{app:repro}

Every GLP is trained by flow matching (Eq.~\ref{eq:fm}) for a single pass over its activation
store (100M activations at both hosts, 500M at ESM-2-8M) in fp32, with AdamW (learning rate
$5\times10^{-5}$, PyTorch-default $\beta$ and weight decay) under a cosine schedule with $1\%$
linear warmup and gradient-norm clipping at $1.0$. The batch is $4{,}096$ activations; the
flow-matching time $t$ is drawn uniformly on $[0,1]$ at a $1{,}000$-step resolution. Only
\texttt{n\_blocks} (depth $3$--$24$) and the derived widths ($d_{\text{model}}{=}2d$,
$d_{\text{mlp}}{=}4d$) vary across the grid. Generation uses $20$ Euler steps; FD uses
$1{,}000$.

\section{On-manifold steering}
\label{app:steering}

\subsection{The steering and decoding procedure}
\label{app:steer-algo}

Algorithm~\ref{alg:steer} states the full procedure; every generation in this paper is one call
to it, and the naive and GLP-corrected arms differ \emph{only} in whether
line~\ref{line:correct} executes. Two details depart from the obvious setup. The intervention is
applied to the \emph{masked} positions only, at a single layer, and re-applied from scratch each
round rather than persisting in the model. Furthermore, the correction operates on the steered activation before decoding, so the language-model head always sees a GLP sample, never the raw steering sum
$h + \alpha v$.

\begin{algorithm}[htbp]
\caption{On-manifold steering by masked-Gibbs decoding, in its property-steering form.
$M$: ESM-2-8M; $h^{(4)}$: its layer-4 residual stream (the tensor GLP is trained on);
$u_\theta$: the GLP velocity field; $S$: the standardizer fitted on the activation store;
$\mathcal{P}(x)$: the amino-acid positions of $x$, cls/eos excluded;
$\overline{\lVert h\rVert}$: the mean activation norm of the property's held-out starting
sequences. The naive and
corrected arms differ \emph{only} in whether line~\ref{line:correct} executes. Canonical
settings: $\rho{=}0.10$, $R{=}30$, $T{=}1.0$, $t_0{=}0.7$, $K{=}20$. The ``down'' arm is
the same call with $-\alpha$, from held-out proteins at the opposite extreme. The
SAE-clamp arm runs the same loop with five changes, listed in
Appendix~\ref{app:steer-sae-clamp}.}
\label{alg:steer}
\begin{algorithmic}[1]
\Require starting sequence $x$, steering vector $v$, strength $\alpha$ (as a fraction of
$\overline{\lVert h\rVert}$), mask fraction $\rho$, rounds $R$, temperature $T$, SDEdit
level $t_0$, Euler steps $K$, flag \textsc{correct}
\State $\hat v \gets \overline{\lVert h\rVert}\,v/\lVert v\rVert$
  \Comment{direction at the scale of a typical activation}
\State $k \gets \max\!\big(1,\ \operatorname{round}(\rho\,|\mathcal{P}(x)|)\big)$
\For{$r = 1 \dots R$}
  \State $P \gets$ $k$ positions drawn uniformly without replacement from $\mathcal{P}(x)$
  \State $x[P] \gets \texttt{<mask>}$
  \State $h \gets h^{(4)}(M, x)$ \Comment{first forward pass, on the masked sequence}
  \For{$p \in P$}
    \State $g \gets h_p + \alpha\,\hat v$ \label{line:steer}
      \Comment{diff-of-means edit}
    \State $\tilde h_p \gets g\,\lVert h_p\rVert / \lVert g\rVert$
      \Comment{norm preserved per residue}
  \EndFor
  \If{\textsc{correct}}
    \State $\tilde h_P \gets \textsc{SDEdit}(\tilde h_P;\, t_0, K)$ \label{line:correct}
    \Comment{project back onto the GLP manifold}
  \EndIf
  \State $\ell \gets M(x)$ with $h^{(4)}_P$ overridden by $\tilde h_P$
  \Comment{second forward pass}
  \State $x[P] \sim \operatorname{softmax}\!\big(\ell_P[\mathcal{A}]/T\big)$,
  \ $\mathcal{A}$ the 20 amino acids
\EndFor
\State \Return $x$
\Statex
\Function{SDEdit}{$a;\, t_0, K$}
  \State $\Delta \gets t_0/K$; \quad $z \gets S(a)$; \quad $\epsilon \sim \mathcal{N}(0, I)$
  \State $z \gets (1-t_0)\,z + t_0\,\epsilon$ \Comment{re-noise to time $t_0$}
  \For{$i = 0 \dots K-1$}
    \State $z \gets z - \Delta\, u_\theta\!\big(z,\ t_0 - i\Delta\big)$
    \Comment{Euler step towards $t{=}0$}
  \EndFor
  \State \Return $S^{-1}(z)$
\EndFunction
\end{algorithmic}
\end{algorithm}

\subsection{Property steering details}
\label{app:steer-estimand}

We report the \textbf{divergence}, the effect of steering up minus the effect of steering down,
tested unpaired (Mann--Whitney) since the two directions start from different held-out
sequences. The two are measured separately and only then subtracted. The divergence also removes the
correction's own $\alpha{=}0$ offset, which is large (Appendix~\ref{app:steer-drift}) and would
otherwise dominate a raw corrected-vs-naive comparison.

In order to make all three properties comparable, charge is length normalized. Net charge is a
sum over residues, so ranking proteins by it ranks them substantially by length, and the
diff-of-means vector would then encode length as much as charge. GRAVY is already a per-residue
mean and the instability index is already length-normalized.

\newpage
Furthermore, \textbf{$\alpha$ is expressed relative to the activation norm.} The raw
diff-of-means vectors have norms $3.70$, $2.94$ and $3.33$, so one grid of $\alpha$ in raw
units would be three different perturbations. We therefore set the direction's norm to
$\overline{\lVert h\rVert}$, the mean of per-residue norms over that property's held-out
starting sequences ($28.83$, $28.28$ and $28.49$ for GRAVY, charge and instability),
$\hat v = \overline{\lVert h\rVert}\,v/\lVert v\rVert$, so that $\alpha$ adds the same norm for
every property. The addition also lengthens the activation, so we rescale the sum back to
$\lVert h\rVert$, $\tilde h = \lVert h\rVert\,(h + \alpha\hat v)/\lVert h + \alpha\hat v\rVert$.
Rescaling does not change direction, so the edit's effect is a rotation of
$\approx\arctan\big(\alpha\,\overline{\lVert h\rVert}/\lVert h\rVert\big)$ at a given residue,
with the norm unchanged. 

An edited activation can fall outside the manifold in two ways: its norm can be different from anything the model produces, or its direction can point to regions where the model never points. The rescaling performed rules out the former to focus on measuring the effects on the latter. The grid is uneven because what the edit adds is $\alpha$ times the steering vector's norm, and
the rescaled vector is $\overline{\lVert h\rVert}/\lVert v\rVert \approx 8$ times longer than
the raw one. In order to keep perturbations unchanged, each raw strength is divided by that factor: $5$, for instance, becomes $\alpha = 0.642$. We sweep
$\alpha \in \{0, 0.032, 0.064, 0.128, 0.257, 0.385, 0.642\}$, and report effects in units of
each property's standard deviation across the $5{,}790$-sequence UniRef50 pool the contrasts
were drawn from ($0.456$ GRAVY, $0.060$ charge/residue, $15.73$ index).

\begin{table}[htbp]
\centering
\caption{Steering-attributable divergence (up $-$ down); $\alpha$ in units of $\overline{\lVert h\rVert}$, SD = divergence over the property's UniRef50 spread. Unpaired Mann--Whitney, $n{=}40$ per direction.}
\label{tab:steer-efficacy}
\begin{tabular}{llrrrrr}
\toprule
property & $\alpha$ & naive & $p$ & GLP-corrected & $p$ & SD \\
\midrule
gravy & 0.032 & $-0.171$ & $0.17$ & $+0.248$ & $0.02$ & $+0.54$ \\
gravy & 0.064 & $+0.006$ & $0.93$ & $+0.551$ & $1{\times}10^{-5}$ & $+1.21$ \\
gravy & 0.128 & $+0.112$ & $0.23$ & $+0.969$ & $7{\times}10^{-10}$ & $+2.12$ \\
gravy & 0.257 & $+0.077$ & $0.44$ & $+1.581$ & $2{\times}10^{-13}$ & $+3.46$ \\
gravy & 0.385 & $+0.546$ & $8{\times}10^{-5}$ & $+2.507$ & $4{\times}10^{-14}$ & $+5.49$ \\
gravy & 0.642 & $+1.400$ & $1{\times}10^{-10}$ & $+3.382$ & $2{\times}10^{-13}$ & $+7.41$ \\
\addlinespace
charge & 0.032 & $-0.006$ & $0.90$ & $+0.011$ & $0.62$ & $+0.18$ \\
charge & 0.064 & $-0.007$ & $0.62$ & $+0.040$ & $4{\times}10^{-3}$ & $+0.67$ \\
charge & 0.128 & $-0.001$ & $0.96$ & $+0.092$ & $5{\times}10^{-9}$ & $+1.54$ \\
charge & 0.257 & $+0.055$ & $0.02$ & $+0.216$ & $2{\times}10^{-14}$ & $+3.60$ \\
charge & 0.385 & $+0.157$ & $6{\times}10^{-10}$ & $+0.313$ & $2{\times}10^{-14}$ & $+5.23$ \\
charge & 0.642 & $+0.430$ & $2{\times}10^{-12}$ & $+0.560$ & $3{\times}10^{-13}$ & $+9.34$ \\
\addlinespace
instability & 0.032 & $-6.17$ & $0.12$ & $+6.77$ & $0.02$ & $+0.43$ \\
instability & 0.064 & $+4.43$ & $0.25$ & $+11.73$ & $2{\times}10^{-3}$ & $+0.75$ \\
instability & 0.128 & $+3.68$ & $0.32$ & $+13.55$ & $3{\times}10^{-5}$ & $+0.86$ \\
instability & 0.257 & $+5.30$ & $0.19$ & $+25.04$ & $7{\times}10^{-9}$ & $+1.59$ \\
instability & 0.385 & $+15.65$ & $2{\times}10^{-4}$ & $+39.48$ & $1{\times}10^{-11}$ & $+2.51$ \\
instability & 0.642 & $+17.01$ & $5{\times}10^{-5}$ & $+64.93$ & $4{\times}10^{-14}$ & $+4.13$ \\
\bottomrule
\end{tabular}
\end{table}

\subsection{The correction's own concept drift}
\label{app:steer-drift}

\begin{figure}[H]
\centering
\includegraphics[width=0.7\textwidth]{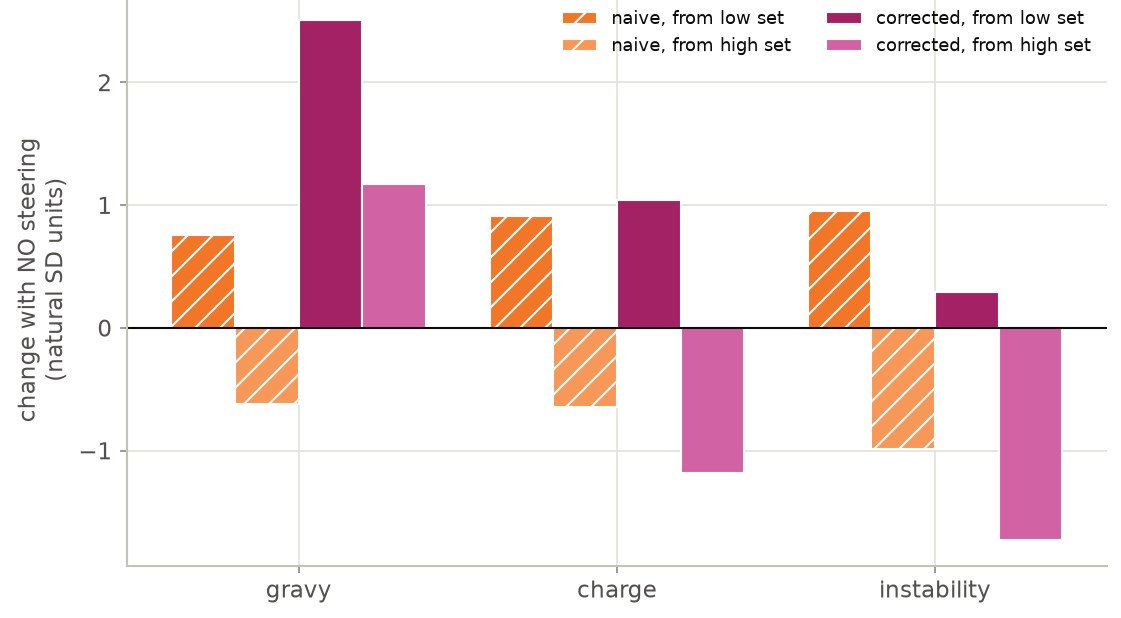}
\caption{The $\alpha{=}0$ arms, in units of each property's spread. Hatched: naive. A bias
that reverses with the starting set is mean reversion; one that does not comes from the
correction.}
\label{fig:steer-drift3}
\end{figure}

The $\alpha{=}0$ run isolates what the loop does without steering. The edit is the identity
there, but the loop still masks and resamples 10\% of residues for 30 rounds, and in the
corrected arm SDEdit still noises the activation to $t_0$ and denoises it back. Both arms move
the property, and what matters is whether the movement is neutral or directional, since an
operator that always pushes one way could be mistaken for a steering effect. The naive arm moves GRAVY $+0.35$ from hydrophilic starts and $-0.28$ from
hydrophobic ones: both move toward the middle, which is resampling reverting to typical
composition. The corrected arm moves $+1.15$ and $+0.54$, both upward, so SDEdit adds
hydrophobicity wherever it starts (Table~\ref{tab:steer-drift}, Fig.~\ref{fig:steer-drift3}). Projecting toward high density is itself a directional move,
toward whatever composition the GLP treats as typical.

\newpage
What the offset is made of is visible in the composition. The same residues move in all three
runs: Leu $+5.9/+6.3/+6.7$pp and Val $+5.4/+5.7/+4.0$pp, with Glu, Lys, Arg and Asp each down
$1.3$--$2.8$pp. The three runs share no data, contrast sets, starting proteins or steering
vectors, only the correction operator. Those
charge losses are also why charge is the only property with no offset: Lys$+$Arg ($4.0$pp) and
Glu$+$Asp ($4.4$pp) are lost in similar amounts, so $\alpha{=}0$ offset on charge is near zero even though the composition changes.

\begin{table}[H]
\centering
\caption{Concept change with \emph{no} steering applied ($\alpha{=}0$), by arm and starting
set, in each property's own units. A fixed bias adds the same amount from both sets; mean
reversion flips sign with the set and cancels in the corrected$-$naive difference, which is
the last column.}
\label{tab:steer-drift}
\begin{tabular}{lrrrrr}
\toprule
& \multicolumn{2}{c}{naive} & \multicolumn{2}{c}{GLP-corrected} & \\
\cmidrule(lr){2-3}\cmidrule(lr){4-5}
property & from low & from high & from low & from high & fixed bias (SD) \\
\midrule
hydrophobicity & $+0.347$ & $-0.282$ & $+1.145$ & $+0.536$ & $+1.77$ \\
charge         & $+0.055$ & $-0.039$ & $+0.063$ & $-0.071$ & $-0.20$ \\
instability    & $+14.99$ & $-15.48$ & $+4.66$  & $-27.13$ & $-0.70$ \\
\bottomrule
\end{tabular}
\end{table}

\begin{figure}[H]
\centering
\includegraphics[width=0.7\textwidth]{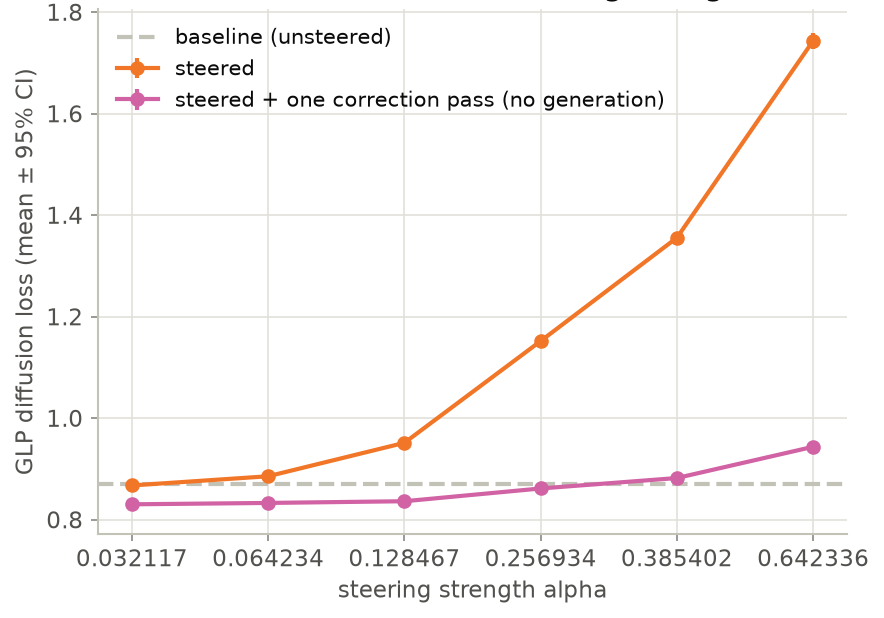}
\caption{GLP diffusion loss against steering strength, before and after a single correction
pass applied directly to the steered activation (no generation), for hydrophobicity}
\label{fig:steer-sdedit}
\end{figure}

\subsection{The correction operator works in isolation}
\label{app:steer-sdedit}

Steering does move activations off the manifold, and monotonically in $\alpha$: GLP's diffusion
loss on real held-out activations rises from $0.869$ unsteered to $1.743$ at the top of the grid
(Fig.~\ref{fig:steer-sdedit}). A single SDEdit pass applied to the steered activation, with
\emph{no} generation, returns it to $0.944$, and recovers over $90\%$ of the excess at every
grid point. What degrades the output, then, is not the edit or the correction but decoding the
corrected activation back to amino acids and iterating that.

\subsection{The decoding protocol, and what it costs}
\label{app:steer-protocol}

\paragraph{The loop has a fixed point.}  Every property contributes two starting sets, one per
steering direction, resulting in six sets: hydrophilic and hydrophobic, anionic and cationic, stable
and unstable. They span natural pTM $0.369$--$0.650$, and the $\alpha{=}0$ naive pass lands every
one of them between $0.227$ and $0.315$ (Table ~\ref{tab:steer-decode6}): the loop does not impose a constant penalty but pulls
everything toward lower pTM values. Nothing beyond the loop adds to that cost. Raising $\alpha$ from zero to the top of the grid barely
moves pTM on any property, neither does the correction on its own, compared against
the naive arm at $\alpha{=}0$. The loop accounts for the foldability lost.

\begin{figure}[H]
\centering
\includegraphics[width=0.8\textwidth]{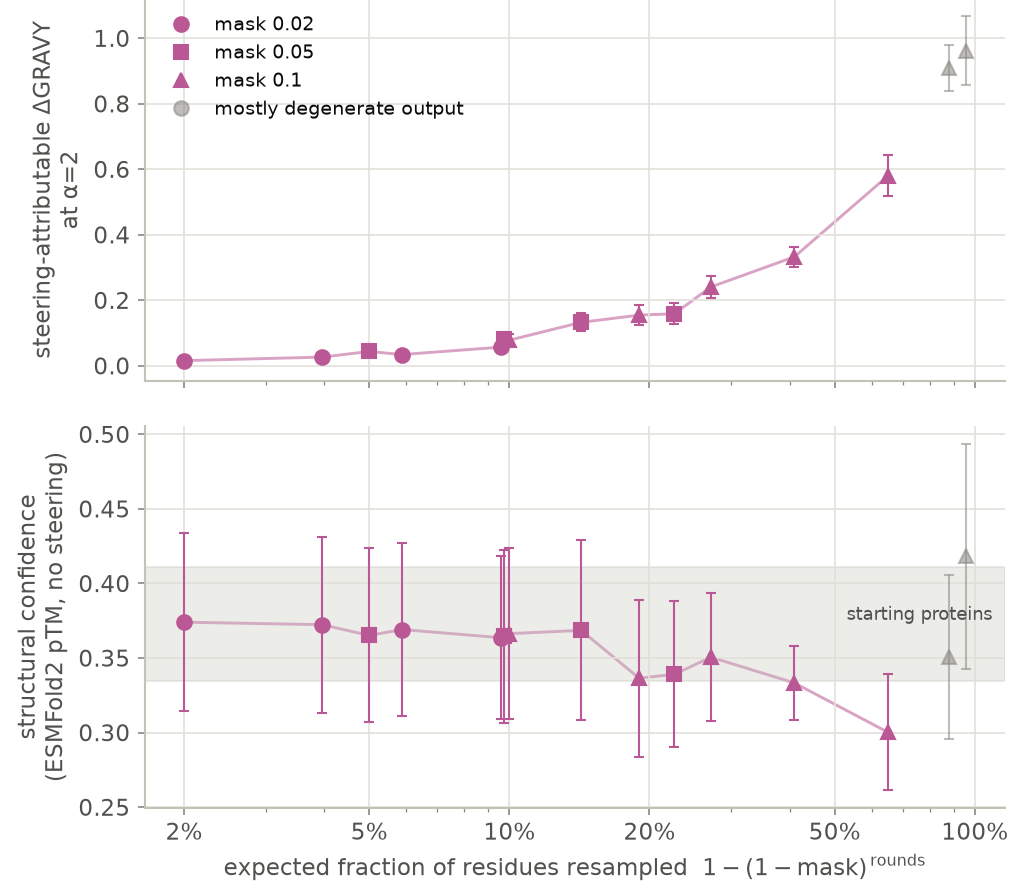}
\caption{Steering effect (top) and foldability (bottom) against how much of the protein the
decoder rewrites, $T{=}0.5$. Marker shape distinguishes mask fraction; the three series fall
on a single curve, i.e.\ mask fraction and round count act only through their product.
Shaded band: the starting proteins' own pTM ($\pm$SEM). Grey points are settings whose
output is mostly degenerate and are excluded from the trend. This budget sweep predates the
relative-$\alpha$ parameterisation and its axis label is in raw units; for the
hydrophobicity vector its $\alpha{=}2$ is the same perturbation as the $\alpha{=}0.257$
used elsewhere in this paper.}
\label{fig:steer-tradeoff}
\end{figure}

\paragraph{Total rewriting is what costs structure.} How the rewriting is split between mask
fraction and rounds does not matter: at matched expected coverage, resampling a small fraction
of residues over many rounds and a larger fraction over few give the same foldability and
comparable effects. What matters is the total, the fraction of residues resampled at least once:
foldability stays within the starting proteins' own interval up to ${\sim}15\%$ coverage and
declines beyond it, while the effect grows throughout (Fig.~\ref{fig:steer-tradeoff}). Steering
strength is not the cost either, since neither raising $\alpha$ across the grid nor the
correction on its own shifts pTM beyond the fixed point.

\begin{table}[htbp]
\centering
\caption{Foldability before and after the decoding loop, with no steering and no correction
($\alpha{=}0$, naive arm), for all six held-out sets.}
\label{tab:steer-decode6}
\begin{tabular}{lrrr}
\toprule
starting set & natural pTM & after decoding & change \\
\midrule
instability, stable         & $0.650$ & $0.294$ & $-55\%$ \\
hydrophobicity, hydrophobic & $0.629$ & $0.315$ & $-50\%$ \\
charge, anionic             & $0.507$ & $0.256$ & $-50\%$ \\
charge, cationic            & $0.438$ & $0.257$ & $-41\%$ \\
hydrophobicity, hydrophilic & $0.373$ & $0.274$ & $-27\%$ \\
instability, unstable       & $0.369$ & $0.227$ & $-38\%$ \\
\bottomrule
\end{tabular}
\end{table}

\subsection{Can a better decoder escape the fixed point?}
\label{app:steer-decoders}

The fixed point of Appendix~\ref{app:steer-protocol} raises whether the foldability loss is specific to uniform-random masked-Gibbs decoding or intrinsic to decoding steered activations at all. We tested two alternative decoders, both using the identical steering vector and correction operator. 

Following the principles of masked generative transformers
\cite{chang2022maskgit}, the first remasks in each round only the lowest-probability positions and freezes the rest. This results in a collapse: $92.7\%$ of outputs are degenerate (${\ge}30\%$
one residue, ${\le}8$ distinct amino acids, or a ${\ge}12$-long homopolymer run) against
$8.5\%$ for uniform random masking, typically homopolymers. For the second, we replace the resampling loop with the stochastic beam search of \cite{beamsearch2026plm}. This approach starts from scores and selects a fixed number of
single-residue substitutions, choosing each one to keep the sequence as likely as possible under
the host. We let the host score those candidates through the steered and corrected activation,
so the search prefers substitutions that move the property while staying plausible. Here, the effect is the opposite: it never collapses and keeps
foldability near natural, but it steers weakly.

\newpage
The two decoders turn out to trace the \emph{same} efficacy/naturalness curve parameterised by
edit count. Beam search makes a fixed number $E$ of single-residue substitutions, and it is that
budget, not the steering strength, that governs the effect: raising $\alpha$ twentyfold barely
changes the effect, whereas growing $E$ increases it steadily while foldability declines
gracefully (Table~\ref{tab:steer-frontier}). Masked-Gibbs rewrites far more than any $E$ we
tested, and extrapolating the beam-search curve to that budget reproduces where it sits: few
edits are natural but weak, many are strong but degrade. The
comparison at $\alpha{=}0$ isolates this: with no steering and no correction, the same starting
proteins fold at pTM $0.48$ through beam search and $0.29$ through masked-Gibbs, so what costs
foldability is the rewriting itself, not the edit applied to the activation. Neither decoder
escapes the trade-off, though beam search is cleaner per edit and may dominate it at matched
effect, a direction we leave open.

\iffalse
\paragraph{Site selection.} We also tried \cite{huang2025diffmean}'s relatedness-ranked site
selection (mutating the residues least aligned with the steering vector instead of a uniform
random subset): at matched edit budget it preserved twice as much of the starting protein but
cost ${\sim}6{\times}$ of the effect ($+0.97{\to}+0.17$) with no significant foldability gain,
plausibly because our single-layer, edited-positions-only intervention is too weak to move the
residues the host model most resists changing (\cite{huang2025diffmean} steer every layer and
token). We retain uniform random masking.

\begin{figure}[t]
\centering
\includegraphics[width=0.52\textwidth]{figures/steer3_decoding_fixed_point.png}
\caption{Foldability of the natural starting proteins against foldability after the decoding
loop with no steering and no correction, for all six held-out sets. The dashed line is no
change; the star marks where the fitted line crosses it. Both steering arms land at this same
mediocre fixed point.}
\label{fig:steer-fp}
\end{figure}
\fi

\begin{table}[H]
\centering
\caption{Beam search over the steered pseudo-likelihood as a function of the edit budget
$E$, against uniform-random masked-Gibbs. pTM is the $\alpha{=}0$ naive arm (the decoder
alone); divergence is the attributable up-minus-down effect at the top of the grid; both at
$n{=}40$ starting proteins per direction, hydrophobicity.}
\label{tab:steer-frontier}
\begin{tabular}{lrrr}
\toprule
decoder & degenerate & pTM ($\alpha{=}0$) & divergence \\
\midrule
beam, $E{=}5$   & $2.5\%$ & $0.486$ & $+0.07$ \\
beam, $E{=}10$  & $2.5\%$ & $0.473$ & $+0.08$ \\
beam, $E{=}25$  & $3.5\%$ & $0.438$ & $+0.27$ \\
beam, $E{=}50$  & $4.6\%$ & $0.409$ & $+0.51$ \\
masked-Gibbs, $30$ rounds & $8.5\%$ & $0.293$ & $+3.99$ \\
\bottomrule
\end{tabular}
\end{table}

\subsection{The SAE-feature-clamp variant: four structural concepts}
\label{app:steer-sae-clamp}

The three biochemical properties are computed from individual residues or neighbouring pairs, which a compositional push can
move, whereas a structural concept depends on the spatial arrangement of specific residues. We therefore repeat the naive/corrected comparison with a
second steering mechanism, clamping one SAE latent ($z_k\!\leftarrow\!c$) as
\cite{villegas2025plmsae} do, on four structural concepts. A clamp of $c{=}1$ sits at the ceiling these features reach on real data; the rest of the grid pushes the latent beyond it. We run this under the same decoder as Algorithm~\ref{alg:steer}, with five changes:
line~\ref{line:steer} becomes the latent clamp; $\rho{=}1$, so each round is an independent
full resample rather than a cumulative refinement; $R{=}100$; $t_0{=}0.5$; finally, the round
maximising $z_k$ is returned, except at $c{=}0$, where there is no clamp and the last round is returned instead. We apply it to Helix (feature 5355),
Beta strand (feature 2868), Coiled coil (feature 813) and Zinc finger (feature 1435, the C2H2 union), each clamped to
the feature with the highest feature--concept F1. Each generation starts from a uniform-random
25--40-residue sequence and re-masks and resamples every position for 100 independent rounds (rather than cumulative, as in property steering), keeping the
round that maximises $z_k$. We test different $c\in\{0,\dots,5\}$ ($\{0,\dots,8\}$ for Zinc finger),
$n{=}100$ per cell.

Unlike the uniform three-property result, the correction's benefit here is concept-dependent
(Table~\ref{tab:steer-sae-clamp}). Helix is the clear case: the correction raises the DSSP helix
fraction on its own, before any clamping, and the naive arm needs a nonzero $c$ to catch
up. Coiled coil appears to gain as well, but inspection shows \texttt{pepcoil} is scoring long
single helices as heptad repeats rather than real pairing. Beta strand barely moves, and the
correction pushes its strand fraction below naive for most $c$. Zinc finger never produces
a complete C2H2 motif, by which we mean a PROSITE hit corroborated by a Pfam zf-C2H2 alignment.
Scoring every sequence continuously against the same profile instead, the corrected arm does
sit higher than naive at most $c$ (Mann--Whitney, best $p{=}2\times10^{-5}$), driven by histidine enrichment rather than motif completion. Fold confidence stays
low throughout.

\textbf{Why helix is the exception.} The clamp sets one scalar per position, a compositional
bias, and the decoder resamples each position from its own distribution, so it can change which
amino acids appear but not how residues sit relative to one another. Local helix propensity
survives that, so helix content rises; a $\beta$-sheet, a coiled coil and a zinc finger each
need specific distant residues paired (two strands hydrogen-bonded, two helices in heptad
register, two Cys and two His chelating an ion), which no per-position bias can arrange.

\textbf{On-manifoldness and naturalness are again compromised.} As with property steering
(Section~\ref{sec:steering-results}), one correction pass returns the clamped activation close
to GLP's typical density in isolation, but that gain is lost at decoding: once the head has
sampled amino acids, ESM re-embeds them and the activation is recomputed from the sequence
rather than carried over, so the finished sequence is no more on-manifold than the naive arm's.

\begin{table}[h]
\centering
\caption{Structural steering by $c$, naive/GLP-corrected, $n{=}100$ per cell.
Structural column: DSSP secondary-structure fraction (per-chain residue fraction in that class)
for Helix and Beta strand; EMBOSS \texttt{pepcoil} hit rate for Coiled coil; and, for Zinc
finger, the fraction of sequences with any PF00096 zf-C2H2 profile-HMM alignment (a graded
measure, no complete motif forms).}
\label{tab:steer-sae-clamp}
\begin{tabular}{rrrrr}
\toprule
$c$ & structure (naive) & structure (corr.) & pLDDT (naive/corr.) & pTM (naive/corr.) \\
\midrule
\multicolumn{5}{l}{\textit{Helix, feature 5355 (DSSP helix fraction)}} \\
0 & 0.54 & 0.77 & 49.7 / 59.8 & 0.153 / 0.251 \\
1 & 0.73 & 0.78 & 59.7 / 62.3 & 0.270 / 0.284 \\
2 & 0.74 & 0.80 & 61.5 / 63.5 & 0.287 / 0.296 \\
3 & 0.76 & 0.77 & 61.8 / 61.7 & 0.290 / 0.285 \\
4 & 0.76 & 0.78 & 61.6 / 62.4 & 0.287 / 0.296 \\
5 & 0.72 & 0.75 & 60.4 / 61.6 & 0.278 / 0.291 \\
\cmidrule(lr){1-5}
\multicolumn{5}{l}{\textit{Beta strand, feature 2868 (DSSP strand fraction)}} \\
0 & 0.001 & 0.002 & 49.7 / 59.8 & 0.152 / 0.251 \\
1 & 0.029 & 0.018 & 49.4 / 52.6 & 0.154 / 0.186 \\
2 & 0.039 & 0.028 & 50.0 / 55.8 & 0.157 / 0.215 \\
3 & 0.033 & 0.045 & 50.3 / 58.1 & 0.163 / 0.235 \\
4 & 0.065 & 0.044 & 50.3 / 59.7 & 0.161 / 0.249 \\
5 & 0.110 & 0.060 & 50.9 / 57.5 & 0.167 / 0.226 \\
\cmidrule(lr){1-5}
\multicolumn{5}{l}{\textit{Coiled coil, feature 813 (\texttt{pepcoil} hit rate)}} \\
0 & 0\% & 0\% & 49.7 / 59.8 & 0.153 / 0.252 \\
1 & 0\% & 0\% & 59.3 / 65.8 & 0.253 / 0.303 \\
2 & 0\% & 4\% & 61.7 / 69.5 & 0.279 / 0.345 \\
3 & 4\% & 15\% & 63.8 / 72.2 & 0.298 / 0.371 \\
4 & 3\% & 19\% & 65.8 / 73.4 & 0.307 / 0.386 \\
5 & 6\% & 33\% & 68.1 / 73.0 & 0.329 / 0.378 \\
\cmidrule(lr){1-5}
\multicolumn{5}{l}{\textit{Zinc finger, feature 1435 (fraction with any PF00096 zf-C2H2 HMM alignment)}} \\
0 & 65\% & 67\% & 49.7 / 59.8 & 0.153 / 0.251 \\
1 & 82\% & 91\% & 49.3 / 54.8 & 0.148 / 0.206 \\
2 & 81\% & 94\% & 51.4 / 56.9 & 0.165 / 0.224 \\
3 & 83\% & 99\% & 50.4 / 51.7 & 0.155 / 0.169 \\
4 & 88\% & 99\% & 51.1 / 48.4 & 0.161 / 0.136 \\
5 & 82\% & 93\% & 51.6 / 47.5 & 0.167 / 0.123 \\
6 & 83\% & 90\% & 53.9 / 48.1 & 0.186 / 0.129 \\
7 & 77\% & 85\% & 57.3 / 47.9 & 0.218 / 0.128 \\
8 & 91\% & 93\% & 59.7 / 48.1 & 0.237 / 0.130 \\
\bottomrule
\end{tabular}
\end{table}

\section{Probing protocol and per-concept tables}
\label{app:tables}

\subsection{Discrete concepts: definition, selection, and probe}
\label{app:probe-discrete}

\textbf{Concepts and splits.} Each discrete concept is one residue-level UniProtKB feature
annotation from Swiss-Prot, materialised per amino acid by InterPLM's
\texttt{extract\_annotations} and named by its feature type and description
(\texttt{Domain\_SH3}, \texttt{Compositional bias\_Acidic residues}, and so on). The evaluation set is a
20{,}000-protein reviewed Swiss-Prot query filtered to proteins with an AlphaFoldDB model and at
most 1{,}022 residues, split by protein into eight shards: 0--3 train the probes, 4--7 are the
held-out test set, both disjoint from GLP's training activations (Appendix~\ref{app:pipeline}).

\textbf{Selection.} We retain every concept that fulfills InterPLM's
frequency filter ($>$25 domain instances or $>$1{,}500 positive residues) and have $\ge$500
positive residues in the test shards. For reference, we tag each concept into one of five biological categories: domain/motif, chemical/PTM, structural, topology/localisation and binding/catalytic. The parsing and filtering yields 110 concepts dominated
by the domain/motif category, which we probe at both hosts.

\newpage
\textbf{Probing.} We probe \emph{computational units} one at a time: one GLP
meta-neuron, one SAE feature, one ESM FFN neuron, or one dimension of the raw residual stream.
The four representations describe the same residues in different coordinate systems, so the
question is which system isolates better a given concept. We therefore run the
following methodology once for each concept in each representation (110 concepts $\times$ 4
representations = 440 probes). First, a unit that detects a concept should be more active on residues that have it than
on residues that do not. In order to account for that, on shards 0--3 we average each unit's activation over the
positive and negative residues, and compute the difference
($\lvert\overline{f}_{\text{pos}}-\overline{f}_{\text{neg}}\rvert$), retaining the 512 units with the largest gap for that concept. Then, we fit each of those as a 1-D $\ell_2$-logistic probe with 5-fold
cross-validation, keeping the one with the highest mean CV (\emph{validation}) AUC, and report its
test AUC on shards 4--7. Negatives come from the same proteins as the positives, so a unit
cannot win on predicting protein identity. We report ROC-AUC, which needs no per-concept threshold and is the metric of GLP's released
probe. InterPLM's headline metric is a domain-adjusted F1, which we have not run on the full
concept set, so we cannot claim the ordering is metric-invariant. GLP's units depend on the diffusion time, so its candidates pool meta-neurons across
$u\in\{0.1,0.5,0.9\}$ and is correspondingly larger than the other three.

\textbf{Falsification.} For the concepts where GLP's selected meta-neuron beats the SAE's
selected feature, we (i) regress it on the full 10{,}240-feature SAE dictionary (held-out
$R^2$: is it a linear read-out?) and (ii) find $k^{*}$, the smallest $k\in\{2,5,10,20\}$ at which a joint probe over $k$ SAE
features matches the test AUC that the single meta-neuron reached. Those $k$ features are the
SAE's highest-ranked candidates by the same positive-minus-negative activation gap used to
shortlist the 1-D probes, computed on train, not by AUC (Table~\ref{tab:app-falsification}).

\begin{figure}[t]
\centering
\includegraphics[width=0.8\textwidth]{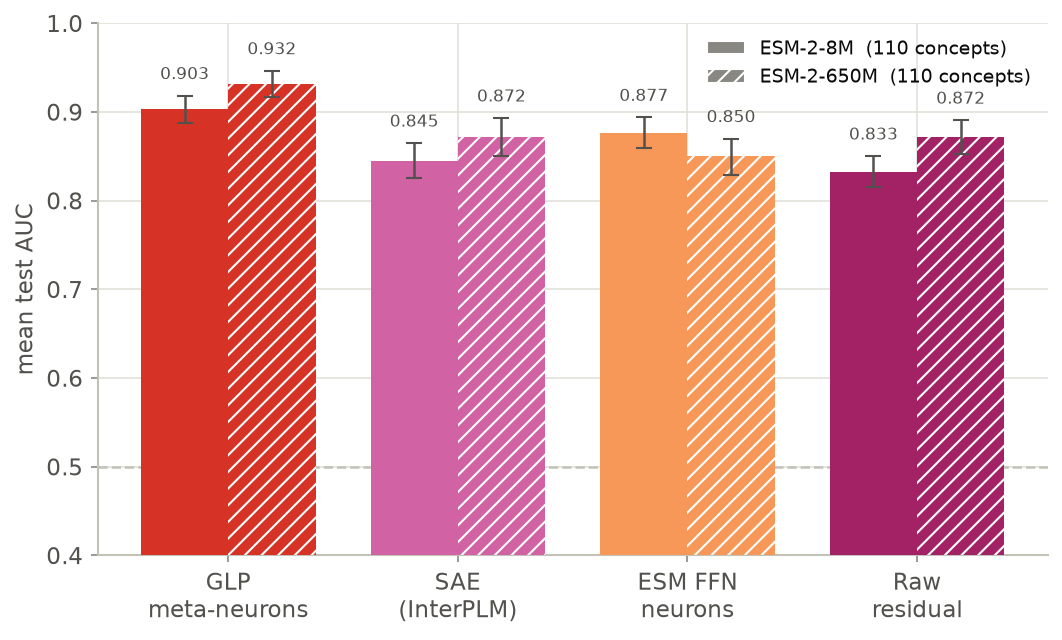}
\caption{1-D probing mean test AUC by method and host (ESM-2-8M solid, ESM-2-650M hatched;
95\% bootstrap CI). GLP meta-neurons lead at both host sizes.}
\label{fig:probing650m}
\end{figure}

\textbf{ESM-2-650M.} The scaling point repeats all of the above on the larger host ($d{=}1280$)
at layer 18, against the released \texttt{Elana/InterPLM-esm2-650m} SAE, with the GLP trained on
100M activations. Because the selection rule depends only on the annotations, the concept set is
identical to the 8M one, so AUCs are comparable across hosts (Fig.~\ref{fig:probing650m}).

\subsection{Continuous properties: definition and probe}
\label{app:probe-continuous}

The ten continuous targets are whole-sequence physicochemical scalars computed directly from the
amino-acid string, with no annotation lookup and no learned predictor. From Biopython's
\texttt{ProteinAnalysis}: GRAVY (mean Kyte--Doolittle hydropathy), molecular weight, isoelectric
point, aromaticity, instability index (Guruprasad dipeptide-weighted sum), net charge at pH 7,
and the helix / turn / sheet secondary-structure fractions. From modlAMP's
\texttt{GlobalDescriptor}: the aliphatic index. Sequences containing a non-canonical residue
(X / B / Z / U / O) are undefined for these scales and dropped.

The probe uses the same shards and the same select-on-validation, report-on-test design as the
discrete one, with three changes forced by the target being a per-protein real number rather
than a per-residue label. First, each protein becomes a single row: a unit's
value for a protein is its activation averaged over that protein's residues. Second,
candidates are ranked by $\lvert$Pearson $r\rvert$ between that pooled value and the target
property, retaining the top $k{=}512$ as well. Third, each 1-D probe is an
ordinary least-squares fit scored by Spearman $\rho$. We select the unit with the largest $\lvert\rho\rvert$ on validation, refit it on shards 0--3, and report its $\rho$ on shards 4--7. We report test Spearman $\rho$ rather
than AUC because a continuous property has no positive / negative split.

\subsection{Per-concept results}
\label{app:probe-tables}

\iffalse
\begin{figure}[t]
\centering
\includegraphics[width=0.3\textwidth]{figures/probing_per_concept.png}
\caption{Per-concept/property GLP vs.\ SAE. Left: discrete concepts, coloured by Swiss-Prot
category; most points sit above the diagonal (GLP wins) across categories rather than
concentrated in one bucket, though the pool is dominated by domain/motif, so a
category-balanced re-run remains future work. Right: continuous properties, direct-labelled;
GLP wins 8/10, led by gravy and sheet fraction.}
\label{fig:probingscatter}
\end{figure}
\fi

Per-concept numbers behind Table~\ref{tab:app-per-concept-both} are presented here. Furthermore,
Table~\ref{tab:app-continuous-both} shows the results for the ten continuous properties and
Table~\ref{tab:app-falsification} the falsification verifications on the concepts GLP wins.

% Generated by scripts/make_probing_tables.py -- do not edit by hand.
% Requires: booktabs, longtable, array, \usepackage[table]{xcolor}
\definecolor{catbg}{HTML}{EDEAF4}
\begingroup\footnotesize\setlength{\tabcolsep}{4pt}
\begin{longtable}{>{\raggedright\arraybackslash}p{4.8cm}rrrrrrrr}
\caption{Per-concept test AUC, all 110 concepts, both hosts. Bold marks the highest value in each row within a host. Shaded rows are category means; * marks a lead over the runner-up with $p{<}0.05$ (two-sided paired Wilcoxon over the category's concepts, uncorrected). Categories with $n{\le}4$ cannot reach this threshold even if the leader wins every concept.} \label{tab:app-per-concept-both} \\
\toprule
 & \multicolumn{4}{c}{ESM-2-8M} & \multicolumn{4}{c}{ESM-2-650M} \\
\cmidrule(lr){2-5}\cmidrule(lr){6-9}
Concept & GLP & SAE & FFN & Res. & GLP & SAE & FFN & Res. \\
\midrule
\endfirsthead
\multicolumn{9}{l}{\textit{(continued)}} \\
\toprule
 & \multicolumn{4}{c}{ESM-2-8M} & \multicolumn{4}{c}{ESM-2-650M} \\
\cmidrule(lr){2-5}\cmidrule(lr){6-9}
Concept & GLP & SAE & FFN & Res. & GLP & SAE & FFN & Res. \\
\midrule
\endhead
\midrule \multicolumn{9}{r}{\textit{continued on next page}} \\
\endfoot
\bottomrule
\endlastfoot
\rowcolor{catbg} \textbf{Domain / motif} \textit{(n=92)} & \textbf{0.904}\rlap{\textsuperscript{*}} & 0.843 & 0.879 & 0.833 & \textbf{0.940}\rlap{\textsuperscript{*}} & 0.879 & 0.867 & 0.895 \\
Compositional bias\_Acidic residues & \textbf{0.974} & 0.972 & 0.920 & 0.920 & \textbf{0.983} & 0.959 & 0.942 & 0.888 \\
Compositional bias\_Basic and acidic residues & \textbf{0.818} & 0.718 & 0.745 & 0.783 & \textbf{0.748} & 0.664 & 0.682 & 0.680 \\
Compositional bias\_Basic residues & \textbf{0.955} & 0.935 & 0.918 & 0.807 & \textbf{0.933} & 0.795 & 0.786 & 0.830 \\
Compositional bias\_Gly residues & \textbf{0.939} & 0.890 & 0.891 & 0.916 & \textbf{0.937} & 0.859 & 0.829 & 0.835 \\
Compositional bias\_Low complexity & \textbf{0.872} & 0.801 & 0.852 & 0.848 & \textbf{0.820} & 0.711 & 0.738 & 0.763 \\
Compositional bias\_Polar residues & 0.856 & 0.809 & \textbf{0.867} & 0.793 & \textbf{0.806} & 0.679 & 0.777 & 0.787 \\
Compositional bias\_Pro residues & \textbf{0.944} & 0.891 & 0.903 & 0.901 & \textbf{0.939} & 0.843 & 0.780 & 0.828 \\
Domain\_C-type lectin & \textbf{0.961} & 0.896 & 0.932 & 0.851 & \textbf{0.986} & 0.938 & 0.904 & 0.973 \\
Domain\_C1q & \textbf{0.950} & 0.832 & 0.898 & 0.834 & \textbf{0.993} & 0.990 & 0.974 & 0.971 \\
Domain\_CTCK & \textbf{0.796} & 0.652 & 0.749 & 0.723 & \textbf{0.977} & 0.940 & 0.913 & 0.881 \\
Domain\_Collagen-like & 0.985 & 0.976 & \textbf{0.988} & 0.976 & \textbf{0.989} & 0.822 & 0.966 & 0.961 \\
Domain\_DHHC & \textbf{0.888} & 0.882 & 0.787 & 0.823 & 0.998 & 0.998 & 0.984 & \textbf{0.999} \\
Domain\_EF-hand 1 & \textbf{0.688} & 0.601 & 0.621 & 0.625 & \textbf{0.857} & 0.740 & 0.741 & 0.742 \\
Domain\_EF-hand 2 & 0.719 & \textbf{0.727} & 0.712 & 0.673 & \textbf{0.771} & 0.668 & 0.655 & 0.726 \\
Domain\_EF-hand 3 & 0.702 & 0.702 & \textbf{0.703} & 0.662 & \textbf{0.822} & 0.699 & 0.740 & 0.714 \\
Domain\_EF-hand 4 & \textbf{0.764} & 0.665 & 0.716 & 0.646 & \textbf{0.877} & 0.875 & 0.729 & 0.773 \\
Domain\_EGF-like & 0.955 & 0.937 & \textbf{0.959} & 0.927 & \textbf{0.894} & 0.870 & 0.779 & 0.729 \\
Domain\_F-box & \textbf{0.995} & 0.967 & 0.957 & 0.906 & \textbf{0.993} & 0.986 & 0.987 & 0.930 \\
Domain\_Fe2OG dioxygenase & \textbf{0.934} & 0.912 & 0.926 & 0.899 & \textbf{0.975} & 0.911 & 0.910 & 0.958 \\
Domain\_G-alpha & \textbf{0.925} & 0.858 & 0.876 & 0.802 & \textbf{0.995} & 0.972 & 0.971 & 0.929 \\
Domain\_GOLD & \textbf{0.951} & 0.783 & 0.902 & 0.873 & \textbf{0.950} & 0.834 & 0.816 & 0.917 \\
Domain\_GST C-terminal & \textbf{0.886} & 0.750 & 0.861 & 0.789 & \textbf{0.979} & 0.974 & 0.934 & 0.938 \\
Domain\_GST N-terminal & \textbf{0.977} & 0.944 & 0.940 & 0.887 & \textbf{0.997} & 0.945 & 0.962 & 0.959 \\
Domain\_Globin & \textbf{0.919} & 0.761 & 0.853 & 0.728 & \textbf{0.995} & 0.989 & 0.976 & 0.940 \\
Domain\_H15 & 0.992 & 0.893 & \textbf{0.994} & 0.950 & \textbf{0.997} & 0.988 & 0.975 & 0.979 \\
Domain\_HTH myb-type 1 & \textbf{0.965} & 0.895 & 0.921 & 0.893 & \textbf{0.951} & 0.922 & 0.884 & 0.922 \\
Domain\_HTH myb-type 2 & \textbf{0.936} & 0.880 & 0.916 & 0.897 & \textbf{0.954} & 0.858 & 0.919 & 0.924 \\
Domain\_IGFBP N-terminal & 0.752 & \textbf{0.764} & 0.757 & 0.726 & 0.911 & \textbf{0.915} & 0.804 & 0.878 \\
Domain\_Ig-like & 0.954 & 0.902 & \textbf{0.959} & 0.872 & \textbf{0.957} & 0.914 & 0.742 & 0.949 \\
Domain\_Ig-like C1-type & \textbf{0.943} & 0.867 & 0.925 & 0.816 & \textbf{0.976} & 0.934 & 0.918 & 0.957 \\
Domain\_Ig-like C2-type & \textbf{0.820} & 0.767 & 0.807 & 0.725 & \textbf{0.830} & 0.741 & 0.750 & 0.751 \\
Domain\_Ig-like C2-type 1 & 0.796 & 0.772 & \textbf{0.799} & 0.732 & \textbf{0.813} & 0.692 & 0.693 & 0.706 \\
Domain\_Ig-like C2-type 2 & \textbf{0.815} & 0.705 & 0.759 & 0.733 & \textbf{0.828} & 0.748 & 0.713 & 0.769 \\
Domain\_Ig-like V-type & \textbf{0.941} & 0.909 & 0.921 & 0.856 & \textbf{0.948} & 0.875 & 0.845 & 0.864 \\
Domain\_J & \textbf{0.983} & 0.954 & 0.977 & 0.902 & \textbf{0.995} & 0.989 & 0.978 & 0.950 \\
Domain\_K-box & \textbf{0.953} & 0.866 & 0.910 & 0.765 & \textbf{0.985} & 0.958 & 0.943 & 0.943 \\
Domain\_LIM zinc-binding 1 & \textbf{0.796} & 0.741 & 0.743 & 0.724 & \textbf{0.832} & 0.782 & 0.628 & 0.768 \\
Domain\_LIM zinc-binding 2 & \textbf{0.774} & 0.723 & 0.755 & 0.711 & \textbf{0.788} & 0.736 & 0.688 & 0.739 \\
Domain\_MADS-box & \textbf{0.924} & 0.882 & 0.890 & 0.840 & \textbf{0.995} & 0.981 & 0.970 & 0.988 \\
Domain\_MARVEL & \textbf{0.953} & 0.808 & 0.926 & 0.859 & \textbf{0.996} & 0.992 & 0.994 & 0.962 \\
Domain\_MPN & \textbf{0.808} & 0.760 & 0.751 & 0.647 & \textbf{0.996} & 0.954 & 0.938 & 0.992 \\
Domain\_N-acetyltransferase & \textbf{0.951} & 0.924 & 0.935 & 0.888 & 0.930 & 0.851 & 0.901 & \textbf{0.940} \\
Domain\_Nudix hydrolase & \textbf{0.936} & 0.917 & 0.916 & 0.828 & 0.890 & 0.731 & 0.806 & \textbf{0.919} \\
Domain\_Orange & \textbf{0.866} & 0.781 & 0.838 & 0.754 & \textbf{0.955} & 0.937 & 0.936 & 0.913 \\
Domain\_PB1 & 0.906 & 0.814 & \textbf{0.947} & 0.914 & \textbf{0.971} & 0.937 & 0.953 & 0.969 \\
Domain\_PDZ & \textbf{0.986} & 0.908 & 0.975 & 0.925 & \textbf{0.976} & 0.903 & 0.817 & 0.956 \\
Domain\_PH & \textbf{0.905} & 0.754 & 0.870 & 0.823 & \textbf{0.988} & 0.961 & 0.923 & 0.955 \\
Domain\_PPIase cyclophilin-type & \textbf{0.919} & 0.748 & 0.904 & 0.882 & \textbf{0.992} & 0.978 & 0.986 & 0.970 \\
Domain\_Peptidase S1 & 0.969 & 0.952 & \textbf{0.987} & 0.925 & \textbf{0.979} & 0.890 & 0.921 & 0.955 \\
Domain\_Protein kinase & 0.960 & 0.934 & \textbf{0.965} & 0.962 & \textbf{0.993} & 0.786 & 0.930 & 0.984 \\
Domain\_RRM & \textbf{0.989} & 0.977 & 0.981 & 0.983 & 0.973 & 0.947 & 0.919 & \textbf{0.986} \\
Domain\_RRM 1 & \textbf{0.858} & 0.816 & 0.828 & 0.804 & \textbf{0.932} & 0.761 & 0.784 & 0.831 \\
Domain\_RRM 2 & \textbf{0.837} & 0.813 & 0.817 & 0.822 & \textbf{0.861} & 0.789 & 0.785 & 0.842 \\
Domain\_SH2 & \textbf{0.834} & 0.668 & 0.786 & 0.746 & \textbf{0.967} & 0.939 & 0.889 & 0.902 \\
Domain\_SH3 & \textbf{0.990} & 0.973 & 0.968 & 0.916 & \textbf{0.996} & 0.948 & 0.939 & 0.951 \\
Domain\_SOCS box & \textbf{0.819} & 0.741 & 0.781 & 0.721 & 0.972 & \textbf{0.987} & 0.980 & 0.936 \\
Domain\_Sm & \textbf{0.892} & 0.792 & 0.887 & 0.883 & \textbf{0.990} & 0.895 & 0.857 & 0.980 \\
Domain\_THD & \textbf{0.939} & 0.768 & 0.918 & 0.811 & \textbf{0.988} & 0.950 & 0.975 & 0.925 \\
Domain\_TSP type-1 & \textbf{0.723} & 0.672 & 0.658 & 0.684 & \textbf{0.966} & 0.922 & 0.836 & 0.877 \\
Domain\_Thioredoxin & \textbf{0.922} & 0.823 & 0.908 & 0.740 & \textbf{0.901} & 0.781 & 0.750 & 0.868 \\
Domain\_Tyrosine-protein phosphatase & \textbf{0.901} & 0.875 & 0.869 & 0.893 & \textbf{0.904} & 0.759 & 0.753 & 0.842 \\
Domain\_UBC core & \textbf{0.845} & 0.718 & 0.809 & 0.689 & 0.938 & 0.903 & 0.895 & \textbf{0.943} \\
Domain\_UPAR/Ly6 & \textbf{0.932} & 0.913 & 0.896 & 0.820 & \textbf{0.931} & 0.862 & 0.836 & 0.850 \\
Domain\_Ubiquitin-like & 0.910 & 0.803 & \textbf{0.911} & 0.773 & \textbf{0.993} & 0.896 & 0.812 & 0.944 \\
Domain\_bHLH & \textbf{0.934} & 0.849 & 0.931 & 0.883 & \textbf{0.990} & 0.978 & 0.988 & 0.984 \\
Domain\_bZIP & \textbf{0.983} & 0.936 & 0.979 & 0.966 & \textbf{0.990} & 0.963 & 0.964 & 0.984 \\
Domain\_sHSP & 0.970 & 0.964 & \textbf{0.973} & 0.956 & \textbf{0.991} & 0.980 & 0.955 & 0.946 \\
Domain\_t-SNARE coiled-coil homology & \textbf{0.917} & 0.845 & 0.867 & 0.806 & 0.987 & 0.994 & 0.967 & \textbf{0.994} \\
Domain\_v-SNARE coiled-coil homology & \textbf{0.967} & 0.946 & 0.939 & 0.910 & \textbf{0.997} & 0.995 & 0.992 & 0.980 \\
Motif\_Effector region & 0.995 & \textbf{0.996} & 0.986 & 0.981 & \textbf{1.000} & 0.993 & 0.992 & 0.993 \\
Motif\_Nuclear localization signal & \textbf{0.852} & 0.838 & 0.802 & 0.735 & \textbf{0.777} & 0.675 & 0.633 & 0.690 \\
Motif\_Nudix box & 0.977 & 0.961 & \textbf{0.978} & 0.943 & \textbf{0.987} & 0.959 & 0.887 & 0.969 \\
Motif\_Switch 1 & \textbf{0.988} & 0.968 & 0.987 & 0.933 & \textbf{0.990} & 0.965 & 0.972 & 0.968 \\
Motif\_Switch 2 & \textbf{0.987} & 0.944 & 0.963 & 0.936 & 0.993 & \textbf{0.997} & 0.965 & 0.941 \\
Region\_Alpha-1 & \textbf{0.887} & 0.769 & 0.860 & 0.796 & \textbf{0.978} & 0.950 & 0.931 & 0.926 \\
Region\_Alpha-2 & \textbf{0.764} & 0.631 & 0.708 & 0.744 & \textbf{0.902} & 0.815 & 0.834 & 0.831 \\
Region\_Alpha-3 & \textbf{0.994} & 0.938 & 0.988 & 0.842 & \textbf{0.999} & 0.993 & 0.959 & 0.991 \\
Region\_Basic motif & \textbf{0.978} & 0.969 & 0.974 & 0.949 & \textbf{0.979} & 0.971 & 0.948 & 0.976 \\
Region\_Disordered & 0.876 & 0.763 & 0.870 & \textbf{0.877} & \textbf{0.856} & 0.682 & 0.778 & 0.767 \\
Region\_Helix-loop-helix motif & \textbf{0.948} & 0.869 & 0.877 & 0.856 & \textbf{0.977} & 0.947 & 0.961 & 0.960 \\
Region\_LID & \textbf{0.881} & 0.775 & 0.807 & 0.816 & \textbf{0.972} & 0.914 & 0.958 & 0.943 \\
Region\_Leucine-zipper & \textbf{0.943} & 0.903 & 0.913 & 0.869 & \textbf{0.904} & 0.870 & 0.875 & 0.873 \\
Region\_NMP & \textbf{0.981} & 0.960 & 0.956 & 0.926 & \textbf{0.998} & 0.996 & 0.974 & 0.983 \\
Region\_Switch-II & \textbf{0.989} & 0.969 & 0.967 & 0.940 & \textbf{0.995} & 0.995 & 0.967 & 0.943 \\
Zinc finger\_C2H2-type & \textbf{0.980} & 0.932 & 0.927 & 0.930 & \textbf{0.971} & 0.758 & 0.852 & 0.858 \\
Zinc finger\_C2H2-type 1 & \textbf{0.888} & 0.811 & 0.876 & 0.810 & \textbf{0.898} & 0.822 & 0.800 & 0.820 \\
Zinc finger\_C2H2-type 2 & 0.867 & 0.812 & \textbf{0.880} & 0.825 & 0.855 & 0.779 & 0.781 & \textbf{0.865} \\
Zinc finger\_C2H2-type 3 & 0.853 & 0.785 & \textbf{0.863} & 0.824 & \textbf{0.878} & 0.770 & 0.739 & 0.834 \\
Zinc finger\_C2H2-type 4 & 0.803 & 0.747 & \textbf{0.837} & 0.766 & \textbf{0.844} & 0.737 & 0.745 & 0.788 \\
Zinc finger\_C4-type & \textbf{0.829} & 0.812 & 0.739 & 0.693 & \textbf{0.914} & 0.773 & 0.738 & 0.771 \\
Zinc finger\_RING-type & \textbf{0.967} & 0.941 & 0.960 & 0.872 & \textbf{0.966} & 0.914 & 0.787 & 0.935 \\
Zinc finger\_any & \textbf{0.937} & 0.881 & 0.936 & 0.849 & 0.916 & 0.700 & 0.747 & \textbf{0.920} \\
\addlinespace[0.9ex] \rowcolor{catbg} \textbf{Chemical / PTM} \textit{(n=9)} & \textbf{0.971} & 0.957 & 0.945 & 0.902 & \textbf{0.972}\rlap{\textsuperscript{*}} & 0.944 & 0.781 & 0.755 \\
Disulfide bond & \textbf{0.997} & 0.970 & 0.977 & 0.912 & \textbf{0.993} & 0.972 & 0.919 & 0.827 \\
Glycosylation\_N-linked (GlcNAc...) asparagine & \textbf{0.989} & 0.985 & 0.985 & 0.925 & \textbf{0.993} & 0.976 & 0.701 & 0.736 \\
Glycosylation\_any & \textbf{0.958} & 0.924 & 0.934 & 0.874 & \textbf{0.964} & 0.927 & 0.729 & 0.697 \\
Lipidation & 0.902 & 0.842 & \textbf{0.919} & 0.890 & \textbf{0.917} & 0.908 & 0.821 & 0.894 \\
Modified residue\_N6-acetyllysine & \textbf{0.958} & 0.958 & 0.950 & 0.916 & \textbf{0.963} & 0.945 & 0.830 & 0.695 \\
Modified residue\_N6-succinyllysine & \textbf{0.977} & 0.977 & 0.968 & 0.925 & \textbf{0.972} & 0.971 & 0.852 & 0.702 \\
Modified residue\_Phosphoserine & 0.976 & \textbf{0.976} & 0.947 & 0.867 & \textbf{0.974} & 0.919 & 0.721 & 0.810 \\
Modified residue\_Phosphothreonine & 0.984 & \textbf{0.986} & 0.835 & 0.898 & \textbf{0.980} & 0.939 & 0.755 & 0.719 \\
Modified residue\_Phosphotyrosine & \textbf{0.993} & 0.991 & 0.991 & 0.915 & \textbf{0.995} & 0.937 & 0.695 & 0.716 \\
\addlinespace[0.9ex] \rowcolor{catbg} \textbf{Structural} \textit{(n=4)} & \textbf{0.752} & 0.656 & 0.694 & 0.646 & \textbf{0.700} & 0.604 & 0.635 & 0.626 \\
Beta strand & \textbf{0.748} & 0.621 & 0.659 & 0.619 & \textbf{0.714} & 0.555 & 0.601 & 0.627 \\
Coiled coil & \textbf{0.878} & 0.829 & 0.827 & 0.762 & \textbf{0.843} & 0.772 & 0.729 & 0.708 \\
Helix & \textbf{0.782} & 0.615 & 0.692 & 0.624 & \textbf{0.644} & 0.553 & 0.638 & 0.601 \\
Turn & \textbf{0.602} & 0.557 & 0.599 & 0.580 & \textbf{0.597} & 0.535 & 0.572 & 0.569 \\
\addlinespace[0.9ex] \rowcolor{catbg} \textbf{Topology / localisation} \textit{(n=3)} & \textbf{0.957} & 0.939 & 0.929 & 0.905 & \textbf{0.973} & 0.956 & 0.940 & 0.923 \\
Transit peptide\_Chloroplast & \textbf{0.961} & 0.951 & 0.945 & 0.885 & \textbf{0.971} & 0.966 & 0.951 & 0.922 \\
Transit peptide\_Mitochondrion & \textbf{0.954} & 0.929 & 0.914 & 0.923 & \textbf{0.974} & 0.950 & 0.931 & 0.924 \\
Transit peptide\_any & \textbf{0.955} & 0.938 & 0.928 & 0.905 & \textbf{0.975} & 0.952 & 0.939 & 0.923 \\
\addlinespace[0.9ex] \rowcolor{catbg} \textbf{Binding / catalytic} \textit{(n=2)} & 0.768 & 0.691 & 0.750 & \textbf{0.769} & \textbf{0.801} & 0.657 & 0.686 & 0.742 \\
Active site\_Proton acceptor & 0.803 & 0.715 & 0.797 & \textbf{0.817} & \textbf{0.827} & 0.694 & 0.708 & 0.766 \\
Active site\_any & \textbf{0.733} & 0.667 & 0.704 & 0.721 & \textbf{0.775} & 0.620 & 0.665 & 0.718 \\
\end{longtable}
\endgroup

% Generated by scripts/make_probing_tables.py -- do not edit by hand.
% Requires: booktabs, longtable, array, \usepackage[table]{xcolor}
\definecolor{catbg}{HTML}{EDEAF4}
\begin{table}[H]
\centering\footnotesize\setlength{\tabcolsep}{4pt}
\caption{Per-property test Spearman $\rho$, all 10 continuous properties, both hosts. The shaded row is the mean, with the Wilcoxon $p$ for the leader against the runner-up.}
\label{tab:app-continuous-both}
\begin{tabular*}{\textwidth}{@{\extracolsep{\fill}}>{\raggedright\arraybackslash}p{4.0cm}rrrrrrrr}
\toprule
 & \multicolumn{4}{c}{ESM-2-8M} & \multicolumn{4}{c}{ESM-2-650M} \\
\cmidrule(lr){2-5}\cmidrule(lr){6-9}
Property & GLP & SAE & FFN & Res. & GLP & SAE & FFN & Res. \\
\midrule
Aliphatic index & \textbf{0.697} & 0.601 & 0.585 & 0.651 & 0.593 & \textbf{0.691} & 0.403 & 0.486 \\
Aromaticity & \textbf{0.765} & 0.717 & 0.522 & 0.465 & 0.631 & \textbf{0.717} & 0.339 & 0.420 \\
GRAVY (hydropathy) & \textbf{0.803} & 0.638 & 0.648 & 0.570 & \textbf{0.684} & 0.562 & 0.465 & 0.401 \\
Helix fraction & \textbf{0.594} & 0.563 & 0.500 & 0.341 & 0.425 & \textbf{0.426} & 0.339 & 0.316 \\
Instability index & 0.575 & \textbf{0.578} & 0.567 & 0.482 & \textbf{0.524} & 0.524 & 0.450 & 0.474 \\
Isoelectric point & 0.575 & 0.511 & \textbf{0.577} & 0.349 & \textbf{0.450} & 0.448 & 0.249 & 0.231 \\
Molecular weight & 0.723 & \textbf{0.739} & 0.448 & 0.439 & 0.610 & \textbf{0.799} & 0.316 & 0.384 \\
Net charge at pH 7 & 0.582 & 0.495 & \textbf{0.585} & 0.339 & 0.428 & \textbf{0.446} & 0.264 & 0.228 \\
Sheet fraction & \textbf{0.764} & 0.610 & 0.656 & 0.646 & \textbf{0.671} & 0.570 & 0.460 & 0.437 \\
Turn fraction & \textbf{0.609} & 0.566 & 0.547 & 0.501 & \textbf{0.532} & 0.511 & 0.444 & 0.356 \\
\bottomrule
\end{tabular*}
\end{table}

% Generated by scripts/make_probing_tables.py -- do not edit by hand.
% Requires: booktabs
\begin{table}[H]
\centering\small
\caption{Falsification tests on the concepts GLP wins. $R^2$: meta-neuron regressed on the full SAE dictionary (held-out). $k^*$: fewest top-$k$ SAE features whose joint probe matches its test AUC (``none'' if no $k\le20$ does).}
\label{tab:app-falsification}
\begin{tabular*}{\linewidth}{@{\extracolsep{\fill}}lll}
\toprule
 & ESM-2-8M & ESM-2-650M \\
\midrule
Concepts tested & 104 of 110 & 105 of 110 \\
$R^2$, median [IQR] & 0.756 [0.706, 0.842] & 0.725 [0.627, 0.799] \\
Concepts at $k^*=2/5/10/20$ & 7/23/36/28 & 14/43/32/12 \\
No $k\le20$ reaches it & 10 & 4 \\
\quad short by, median & 0.0035 & 0.0041 \\
\quad largest gap & Active site\_Proton acceptor (0.171) & Domain\_RRM 1 (0.054) \\
\midrule
\textit{Examples} & \multicolumn{2}{l}{\textit{$R^2$ / $k^*$ at each host}} \\
\quad Transit peptide\_Chloroplast & 0.880 / 2 & 0.839 / 2 \\
\quad Domain\_Protein kinase & 0.893 / 5 & 0.884 / 10 \\
\quad Region\_Disordered & 0.829 / 10 & 0.903 / 10 \\
\quad Beta strand & 0.770 / 20 & 0.762 / 20 \\
\quad Domain\_SH2 & 0.169 / 10 & 0.523 / 5 \\
\quad Disulfide bond & 0.958 / none & 0.915 / 2 \\
\bottomrule
\end{tabular*}
\end{table}

\end{document}